\documentclass[sigconf]{acmart}
\AtBeginDocument{%
  }

\acmConference[MM '25]{Proceedings of the 2025 ACM Multimedia Conference}{October 27--31, 2025}{Dublin, Ireland}

\usepackage{pifont}
\newcommand{\cmark}{\ding{51}}
\newcommand{\xmark}{\ding{55}}
\usepackage{colortbl}
\usepackage{xcolor}

\definecolor{lightblue}{HTML}{E6FAFD}  
\definecolor{lighterblue}{HTML}{E7EEFC}  
\definecolor{purple}{HTML}{BEE6FC} 
\definecolor{darkerblue}{HTML}{C3D6F2} 

\usepackage{multirow}
\usepackage{CJKutf8}

\begin{document}

\title{Rethinking Multilingual Vision-Language Translation: Dataset, Evaluation, and Adaptation}

\author{%
\textbf{Xintong Wang}$^{1,2}$,
\textbf{Jingheng Pan}$^{2}$,
\textbf{Yixiao Liu}$^{3}$,
\textbf{Xiaohu Zhao}$^{1}$,
\textbf{Chenyang Lyu}$^{1}$, \\
\textbf{Minghao Wu}$^{1,4}$,
\textbf{Chris Biemann}$^{2}$,
\textbf{Longyue Wang}$^{1}$,
\textbf{Linlong Xu}$^{1}$,
\textbf{Weihua Luo}$^{1}$,
\textbf{Kaifu Zhang}$^{1}$
}

\affiliation{%
  \institution{$^{1}$Alibaba International Digital Commerce Group \quad
               $^{2}$Universität Hamburg, Department of Informatics \quad
               $^{3}$Nanyang Technological University \quad
               $^{4}$Monash University}
  \country{}      
}

\email{wanglongyue.wly@alibaba-inc.com}

\renewcommand{\shortauthors}{Wang et al.}

\begin{abstract}
Vision-Language Translation (VLT) is a challenging task that requires accurately recognizing multilingual text embedded in images and translating it into the target language with the support of visual context. While recent Large Vision-Language Models (LVLMs) have demonstrated strong multilingual and visual understanding capabilities, there is a lack of systematic evaluation and understanding of their performance on VLT. In this work, we present a comprehensive study of VLT from three key perspectives: data quality, model architecture, and evaluation metrics. (1) We identify critical limitations in existing datasets, particularly in semantic and cultural fidelity, and introduce AibTrans—a multilingual, parallel, human-verified dataset with OCR-corrected annotations. (2) We benchmark 11 commercial LVLMs/LLMs and 6 state-of-the-art open-source models across end-to-end and cascaded architectures, revealing their OCR dependency and contrasting generation versus reasoning behaviors. (3) We propose Density-Aware Evaluation to address metric reliability issues under varying contextual complexity, introducing the DA Score as a more robust measure of translation quality. Building upon these findings, we establish a new evaluation benchmark for VLT. Notably, we observe that fine-tuning LVLMs on high-resource language pairs degrades cross-lingual performance, and we propose a balanced multilingual fine-tuning strategy that effectively adapts LVLMs to VLT without sacrificing their generalization ability.
\end{abstract}

\maketitle

\section{Introduction}
Vision-Language Translation (VLT) involves translating multilingual text embedded directly within images into a target language \cite{li202410m, lan2023exploring, qian2024anytrans, salesky2024benchmarking}. Unlike multimodal machine translation\cite{khan2025chitranuvad}—where text is externally provided and image serves only as auxiliary context—or image captioning\cite{li2025underwater}—where textual output is generated freely from visual input—VLT requires accurate recognition of in-image text and contextually grounded translation into the target language. This dual challenge of OCR and translation, tightly coupled with visual semantics, makes VLT particularly demanding.

Recent advances in large vision-language models (LVLMs) \cite{bai2025qwen2, li2024llava, chen2024expanding, wu2024deepseek} have brought together the multilingual strength of large language models (LLMs) \cite{yang2024qwen2, bai2023qwen, dong2024internlm} and the ability to natively interpret visual content. Enabled by ongoing progress in generative and reasoning capabilities \cite{guo2025deepseek}, LVLMs have demonstrated strong performance across a range of vision-language tasks, including the challenging VLT task. However, despite this progress, the field still lacks evaluation frameworks that match the scale and complexity of these models. In particular, there is limited understanding of the performance boundaries, behavioral characteristics, and cross-lingual translation capabilities of modern LVLMs—especially when applied to diverse and visually grounded translation scenarios.

In this work, we revisit the VLT task by systematically analyzing data quality, model architecture, evaluation protocols, and fine-tuning strategies. Our findings reveal surprising gaps and underexplored strengths in current approaches. First, we observe that widely used VLT datasets suffer from reliability issues, especially in their OCR and reference translations. Despite being labeled as human-annotated, many reference sentences are grammatically well-formed but semantically flat or culturally misaligned—an issue observed not only in low-resource languages but also in high-resource ones like Chinese. 


We then explore how different model architectures behave under the VLT setting. End-to-end models that integrate OCR and translation demonstrate better average performance by leveraging visual grounding, while cascaded models with external OCR modules exhibit more stable behavior across datasets. Among open-source models, Qwen2.5-VL \cite{bai2025qwen2} performs competitively with state-of-the-art commercial systems. We also examine the growing class of reasoning-based models, which refine translation outputs through multi-step reasoning. Interestingly, we find that using strict task-specific instructions—commonly employed for generation models—can degrade performance in reasoning models by triggering overthinking and deviation from contextually optimal solutions.

To further understand model performance, we analyze evaluation metrics across diverse translation scenarios. Our results show that common metrics such as BLEU, BERTScore, and COMET behave inconsistently depending on contextual complexity, especially sentence length and the number of recognized text regions. This variability leads to misaligned estimates of model quality when scores are naively averaged.

In response, we contribute AibTrans, a high-quality, multilingual, parallel VLT dataset with manually corrected OCR and culturally faithful translations spanning seven target languages. To account for metric inconsistency, we propose a Density-Aware Scoring (DA Score) scheme based on information density—defined jointly by bounding box count and token length—offering a context-sensitive view of model performance. Finally, we propose a balanced multilingual fine-tuning strategy that adapts LVLMs to the VLT task while preserving their generalization capabilities. Instead of fine-tuning on a single high-resource language pair, we sample 1000 balanced examples across multiple directions. This improves task alignment without degrading multilingual performance and highlights the importance of data diversity in VLT adaptation. Through these contributions, we establish a unified benchmark for evaluating multilingual vision-language translation and provide actionable insights for future research at the intersection of vision, language, and multilingual.

\section{Data Landscape and Challenges}

\subsection{Existing Datasets and Their Limitations}
Recently released datasets such as MIT-10M \cite{li202410m}, OCRMT30K \cite{lan2023exploring}, and MTIT6 \cite{qian2024anytrans} have facilitated the development and evaluation of VLT tasks by offering large-scale, multilingual resources grounded in real-world visual scenarios. However, these datasets differ significantly in construction and quality, which can be examined along three key factors. \textbf{(1) OCR reliability and correction.} OCR errors are a major source of noise in VLT datasets. MIT-10M uses EasyOCR \cite{jaided2020easyocr} for initial recognition, followed by GPT-4o \cite{openai2025gpt4o} refinement. However, no manual correction is performed, and GPT-4o may hallucinate text in visually ambiguous regions. OCRMT30K and MTIT6 use PaddleOCR without post-correction, resulting in recognition errors being silently propagated into the reference translations. \textbf{(2) Use of visual context during translation.} In MIT-10M, reference translations are generated by GPT-4 \cite{openai2023gpt4} using only the OCR output, without access to image content. In contrast, OCRMT30K and MTIT6 involve human translators with access to the original images. Nonetheless, due to the uncorrected OCR stage, visual errors may still influence the final translations, limiting the benefits of image grounding. \textbf{(3) Language coverage and parallelism.} MIT-10M covers eight source languages, including high-resource languages such as English and Chinese and low-resource languages such as Hindi, Turkish, and Arabic. Each is translated into thirteen target languages, forming a fully parallel multilingual dataset. OCRMT30K includes only Chinese-to-English translations. MTIT6 supports six translation directions, but lacks parallel alignment across languages. Additionally, MIT-10M and OCRMT30K provide both training and evaluation splits, while MTIT6 is designed for evaluation only.

\subsection{Semantic and Cultural Concerns}
While we analyzed the construction process of existing VLT datasets, a key question remains: \textbf{whether the provided reference translations offer reliable evaluation targets}. To address this, we assess the translation quality of the references in each dataset.

We use two strong multilingual LLMs, Qwen-Max (commercial) \cite{qwen_max} and DeepSeek-V3 (open-source) \cite{liu2024deepseek}, to evaluate the references along four dimensions: \textbf{Semantic Adequacy}: whether the translation accurately conveys the meaning of the source sentence. \textbf{Grammatical Correctness}: whether the translation conforms to the grammatical rules of the target language. \textbf{Fluency}: whether the translation reads smoothly and naturally to native speakers. \textbf{Cultural Appropriateness}: whether the translation aligns with cultural norms and avoids misunderstandings or offensive expressions.

For each translation, the model receives the source sentence and the corresponding reference translation, and is prompted to assign a score from 1 to 5 for each dimension, with a brief explanation. The overall score is computed as the average of the four dimensions and is used to represent the general reference quality of each dataset.

\begin{figure}[h]
  \centering
  \includegraphics[width=0.8\linewidth]{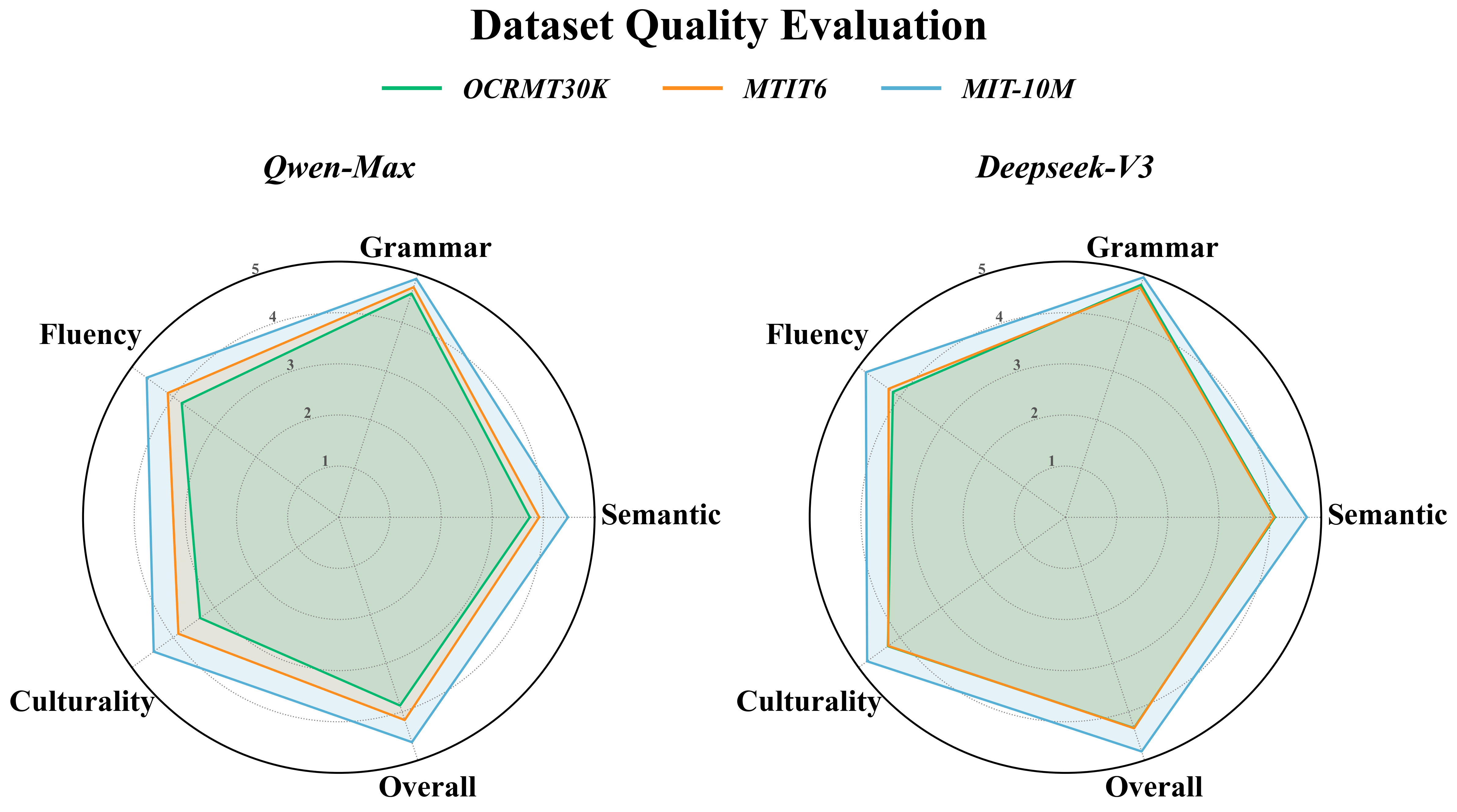}
  \caption{Dataset Quality Evaluation across grammar, semantics, fluency, cultural fit, and overall score.}
  \label{fig:dual_radar_chart} 
\end{figure}

Figure~\ref{fig:dual_radar_chart} reveals clear quality disparities among the three datasets. MIT-10M achieves the highest overall ratings from both Qwen2.5-Max (4.63) and DeepSeek-V3 (4.82), particularly in grammaticality and fluency (all above 4.6). However, semantic adequacy and cultural appropriateness receive slightly lower scores—around 4.48 and 4.47 for Qwen—suggesting that even with GPT-4’s strong generation capabilities, machine-generated translations still face challenges in fully capturing image-grounded meaning and producing culturally nuanced outputs. By contrast, OCRMT30K and MTIT6 exhibit significantly lower reference quality than MIT-10M. Qwen assigns overall scores of 3.87 and 4.17, respectively, with particularly low scores in semantic adequacy (3.73) and cultural fit (3.35) for OCRMT30K. Despite grammatical correctness remaining high (above 4.6), these results suggest that references in these datasets were likely derived from post-edited machine translations. Even when annotators have access to images, they often retain literal or rigid phrasings from the MT output, limiting the incorporation of contextual or culturally nuanced rephrasing. This results in translations that may be grammatically well-formed but are semantically shallow, stylistically unnatural, or culturally out-of-place.

These findings raise a fundamental concern regarding the use of such reference translations for training or evaluation. When grammatical correctness dominates the reference signal—while semantic fidelity and cultural appropriateness remain weak—models may be systematically rewarded for surface-level fluency rather than deeper understanding. As a result, models evaluated against these references may learn to avoid context-sensitive paraphrasing, culturally grounded rephrasings, or semantically richer alternatives, simply because such outputs diverge from the flawed reference. This introduces a form of evaluation bias that punishes creativity, penalizes semantic correctness, and discourages cultural sensitivity.


\begin{figure}[h]
  \centering
  \includegraphics[width=0.65\linewidth]{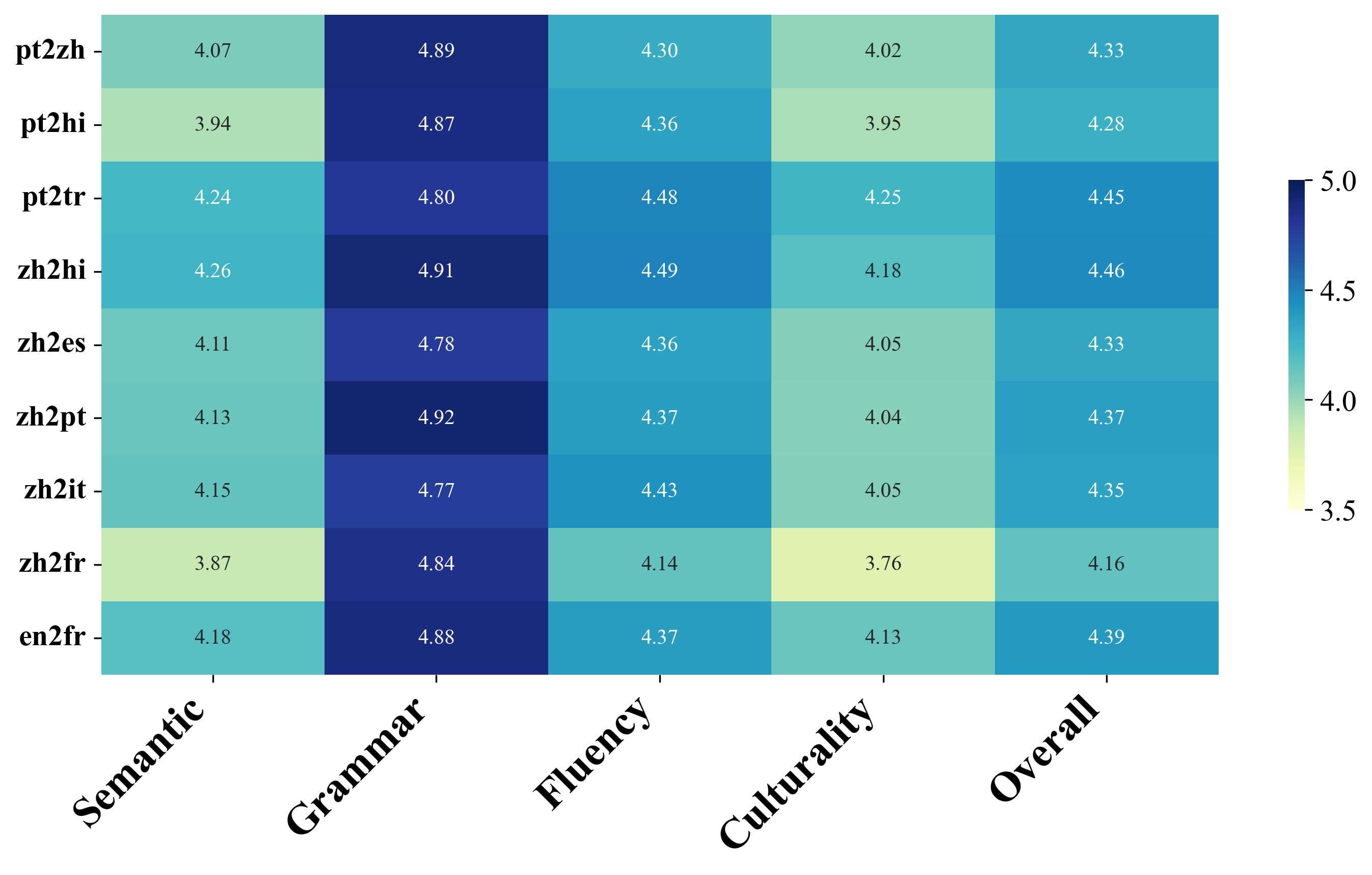}
  \caption{Multilingual Multi-direction Translation Quality Evaluation under dimensions with scores $<$4.3.}
  \label{fig:qwen_heatmap_single} 
\end{figure}

To further investigate the quality of translations in MIT-10M, we conducted a fine-grained analysis across all language pairs. For each source–target direction, we selected cases where any evaluation dimension scored below 4.3. Notably, in Figure~\ref{fig:qwen_heatmap_single}, low semantic or cultural scores are most prevalent in translations from Chinese (5 instances) and Portuguese (3) followed by Spanish (1 instances). This observation is surprising, especially in the case of Chinese, a high-resource language. We hypothesize that the mismatch between surface-level fluency and deeper semantic or cultural fidelity is particularly pronounced in Chinese due to its unique logographic structure and rich socio-cultural expressions.

\subsection{AibTrans: Multilingual Parallel Dataset}
To remedy the limitations observed in existing VLT datasets, we introduce \textbf{AibTrans}, a high-quality, multilingual evaluation dataset built via a two-stage process: OCR correction and human translation.

\begin{figure}[h]
  \centering
  \includegraphics[width=0.8\linewidth]{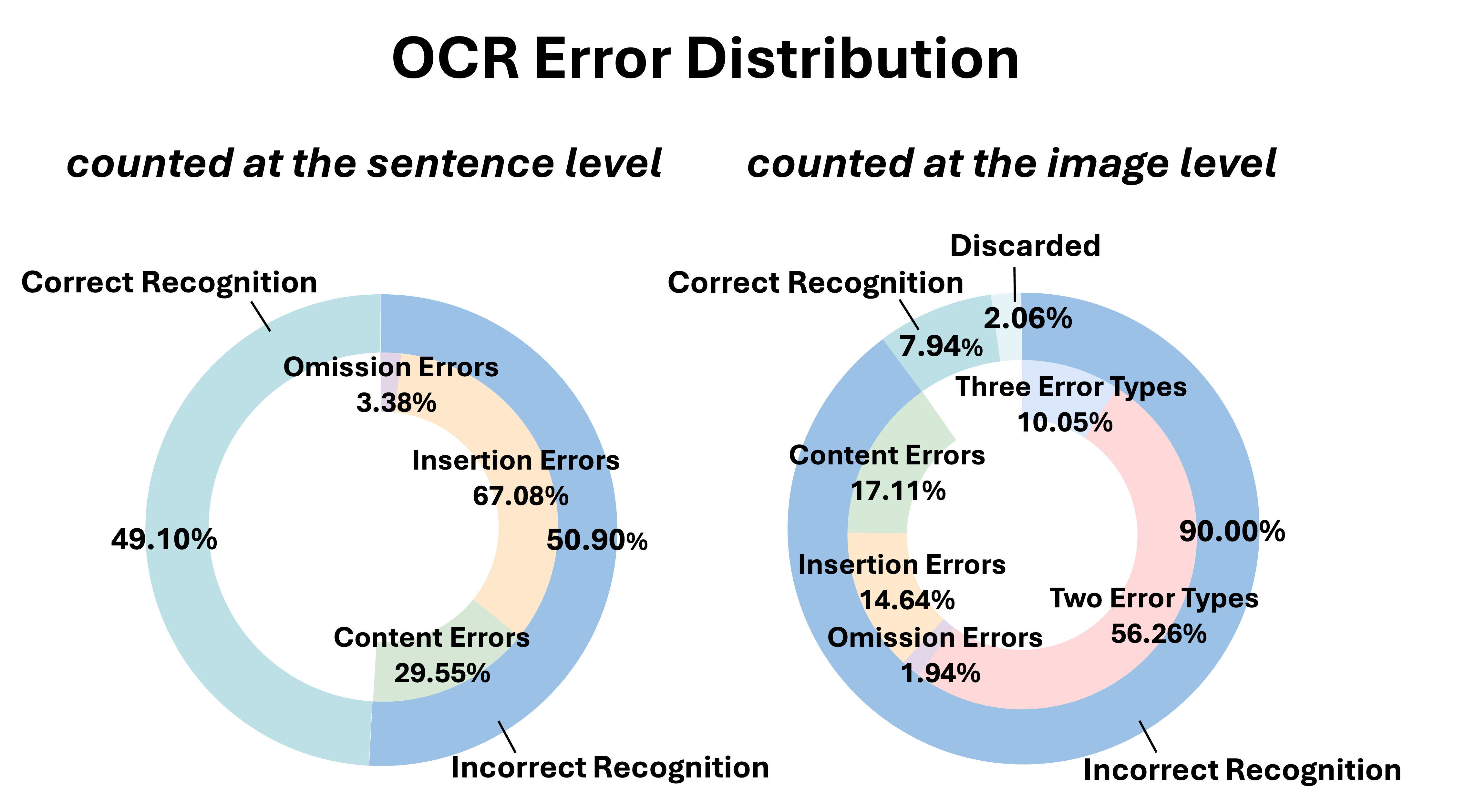}
  \caption{OCR Recognition Error Distribution Analysis}
  \label{fig:ocr_error} 
\end{figure}
In the first step, we collected 2,338 images from Taobao, covering a wide range of challenging visual-textual conditions, including mixed-language text, stylized fonts, varied text layouts, and complex image backgrounds. After manually filtering for diversity and VLT relevance, we selected 630 images for annotation. We first applied PaddleOCR \cite{paddleocr2025} for initial text detection, followed by manual correction. As shown in Figure ~\ref{fig:ocr_error}, only 49.1\% of sentences were error-free, while the remaining 50.9\% required edits. The majority of errors were insertion errors (67.1\%), and 29.6\% contained semantic content issues requiring rewriting. At the image level, only 7.9\% of images were fully accurate, while 90\% contained at least one error. Among erroneous images, 17.1\% had content errors, 14.6\% had insertion errors, and 56.3\% had both. 10.1\% exhibited all three error types. This analysis highlights the necessity of OCR post-correction to ensure data reliability in VLT benchmarks. In the second step, we submitted the corrected OCR results and original images to professional translators for human translation from Chinese into seven target languages: English, German, Spanish, Arabic, Russian, Japanese, and Hindi. These languages were selected to cover high-resource (e.g., English), low-resource (e.g., Hindi), diverse regions (Asia, Europe, Americas, Africa), and translation challenges involving cultural adaptation (e.g., Arabic). Translators were instructed to ground translations in the visual context and to consider cultural appropriateness when choosing lexical and syntactic structures. Following initial translation, a second translation team performed quality checks.

The final dataset comprises parallel translations from Chinese into seven languages, grounded in images and corrected OCR, with a strong emphasis on semantic fidelity, cultural appropriateness, and OCR reliability. As a parallel resource, AibTrans enables fair cross-lingual evaluation and supports fine-grained analysis of multilingual VLT performance across diverse linguistic and cultural conditions.

\begin{figure}[h]
  \centering
  \includegraphics[width=0.9\linewidth]{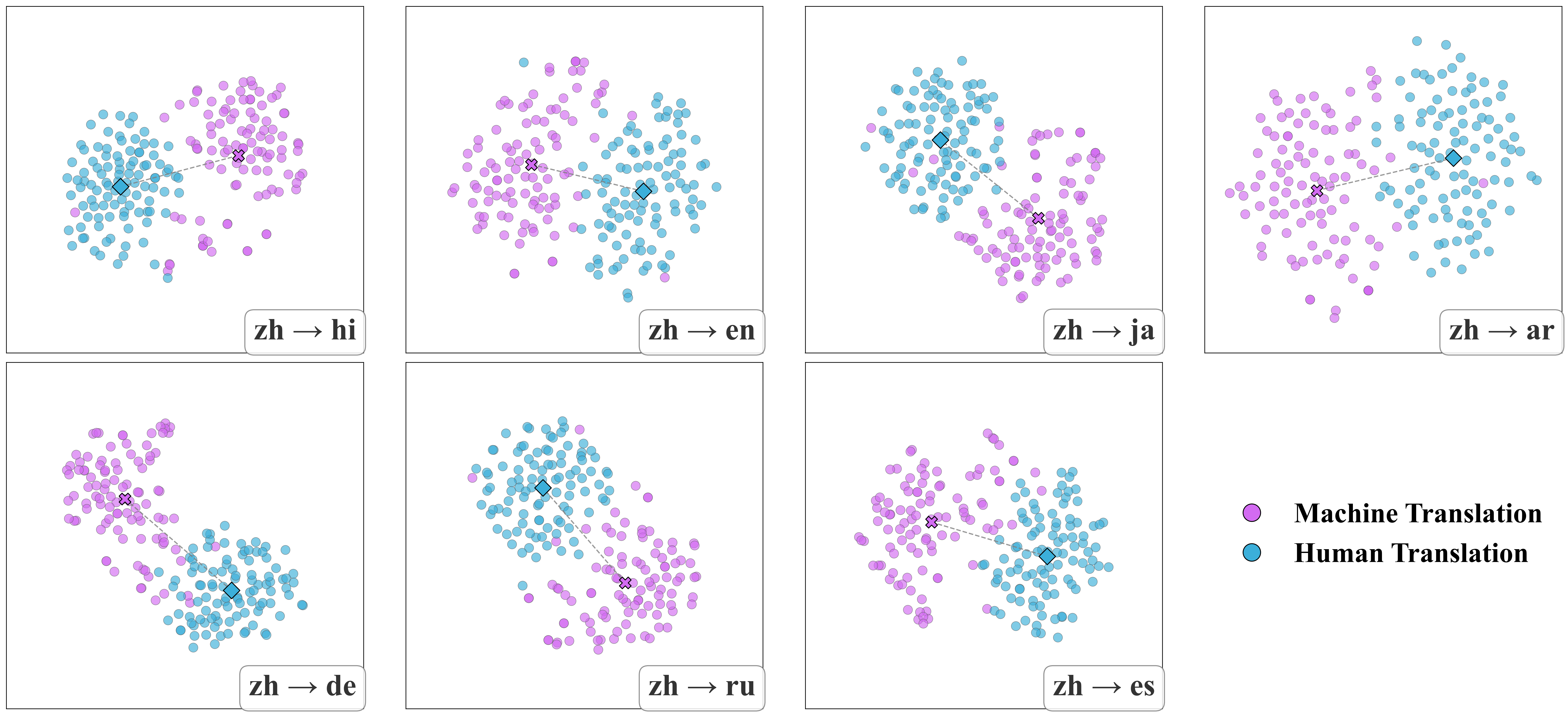}
  \caption{Distributions of Machine vs. Human Translations. Embeddings are from OpenAI embeddings \cite{openai_embeddings_api}}.
  \label{fig:tsne_pic} 
\end{figure}

To further illustrate the distinction between human and machine-generated references, we visualize sentence embeddings of AibTrans (human-translated) and MIT-10M (GPT-4 translated) using t-SNE \cite{van2008visualizing}. As shown in Figure ~\ref{fig:tsne_pic}, the two types of translations form clearly separable clusters, indicating distinct stylistic patterns. Despite the high quality of GPT-4 outputs, their distribution differs significantly from human translations. Relying on such references for evaluation risks overfitting models to machine-like generation patterns, rather than encouraging alignment with authentic human translation styles.
\section{Architectures for VLT}
\subsection{End-to-End vs. Cascaded}

\begin{figure}[h]
  \centering
  \includegraphics[width=0.8\linewidth]{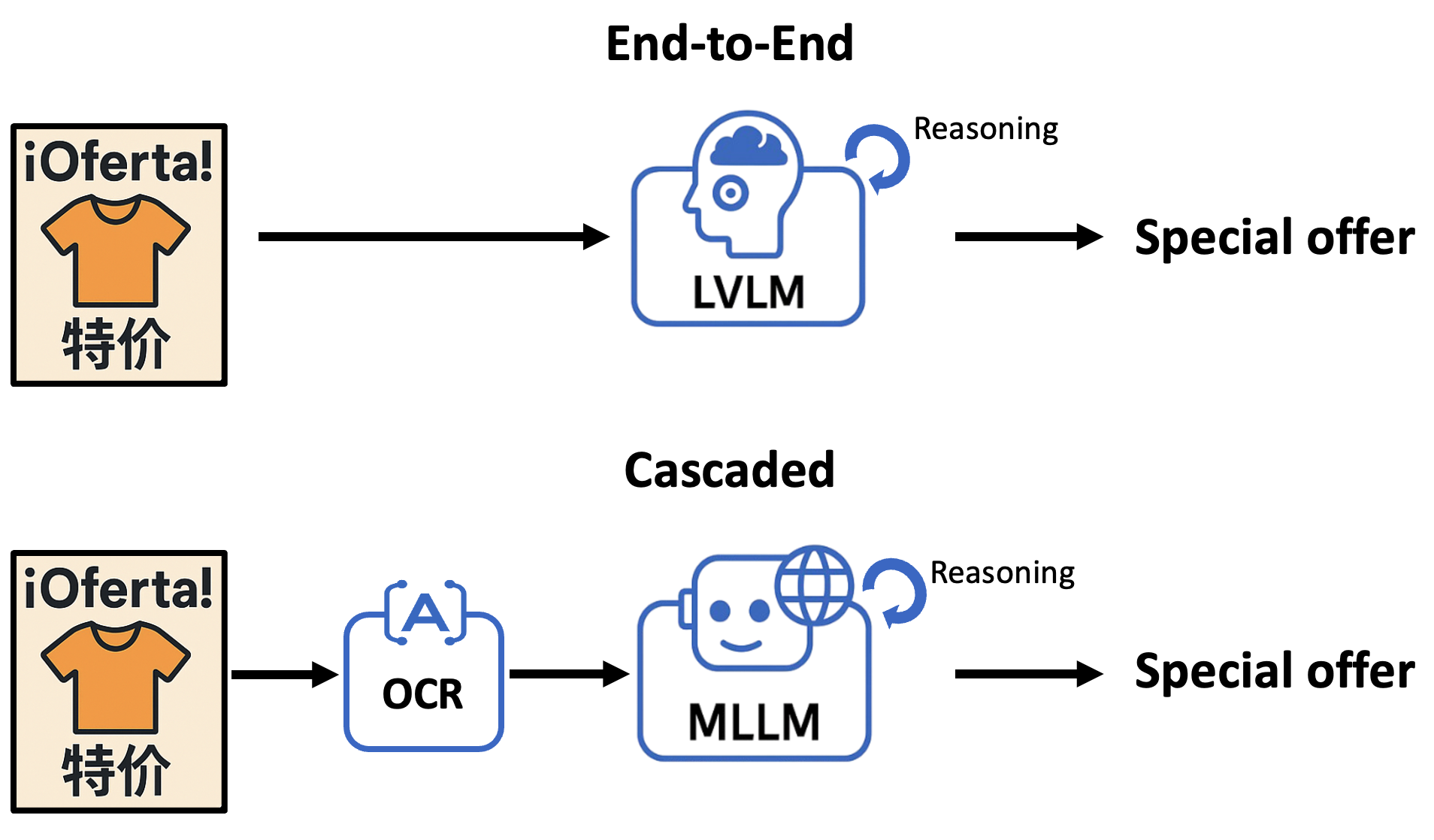}
  \caption{Architectures of End-to-End and Cascaded Model}
  \label{fig:arc} 
\end{figure}

Current approaches generally fall into two architectural paradigms: end-to-end and cascaded. As illustrated in Figure 5, end-to-end models (top) employ LVLMs that directly process images as input. These models implicitly perform OCR while simultaneously generating translations, leveraging visual cues such as layout, typography, and surrounding objects. Cascaded models (bottom) decompose the task into two stages. An OCR tool first extracts the textual content from the image, which is then translated by a multilingual LLM. This setup offers greater modularity and flexibility across languages. Additionally, the translation model can sometimes correct OCR errors—especially obvious insertions or malformed tokens—based on language priors. Nevertheless, performance is still tightly coupled with OCR quality; failures such as over- or under-recognition can adversely affect the final output.



Both architectures may use either generative or reasoning-based models, as denoted by the circular loop in Figure ~\ref{fig:arc}. Generative models translate text in a single forward pass, while reasoning-based models engage in multi-step reasoning, explicitly leveraging OCR outputs or visual features before producing the final translation.

\subsection{Quantitative Results across Architectures}

To assess the translation capabilities of different architectures in VLT, we conduct a comprehensive evaluation across three representative datasets: MIT-10M, OCRMT30K, and MTIT6. We report BLEU, BERTScore (BS-F1), and COMET as our main evaluation metrics, covering both n-gram overlap and sentence-level semantic similarity. BLEU captures exact lexical match, while BERTScore and COMET provide deeper semantic evaluations. We evaluate two types of models: commercial closed-source systems and open-source alternatives. For commercial models, we evaluate GPT-4o, Gemini-2.0-flash \cite{google2025gemini2flash}, Qwen-VL-Max \cite{qwen2024vlmax}, Claude3.7 \cite{anthropic2025claude37}, GPT4-o1 \cite{openai2025gpt4o1}, and QvQ-72B-preview \cite{qwen2025qvq72b}. For cascaded commercial setups, we include Deepseek-v3, Macro-MT \cite{macro2025mt}, Qwen2.5-Max, Deepseek-r1 \cite{guo2025deepseek}, and QwQ-32B-preview \cite{qwen2024qwq32b}. Specifically, GPT4-o1, QvQ-72B-preview, Deepseek-r1, and QwQ-32B-preview function as reasoning models, whereas the remaining systems operate as generation models. For open-source models, we experiment with state-of-the-art end-to-end models including Qwen2.5-VL (7B/3B) \cite{bai2025qwen2}, InternVL2.5 (8B/2B) \cite{chen2024expanding}, and LLaVA-OneVision-7B \cite{li2024llava}, as well as their cascaded counterparts using identical multilingual LLMs such as Qwen2.5-7B/3B-Instruct \cite{yang2024qwen2}, InternLM2.5-7B/1.8B-Chat \cite{dong2024internlm}, and Qwen2-7B-Instruct \cite{bai2023qwen}. This setup allows us to establish reference ceilings with commercial models and analyze how well open-source solutions approach these results under comparable conditions

In Table~\ref{tab:eval_3dataset}, end-to-end commercial models like GPT-4o (58.82), Qwen-VL-Max (59.35), and GPT4-o1 (58.29) slightly outperform cascaded ones, benefiting from direct visual grounding and strong internal OCR. Deepseek-v3 (58.07) stands out among cascaded models, demonstrating the effectiveness of separating OCR and translation. While cascaded models show stable performance (53.83–58.07), end-to-end models vary more widely (49.60–59.35), likely due to implicit OCR failures or hallucinations within their unified architecture. OCR insertion errors, although frequent, mainly affect BLEU and have less impact on sentence-level metrics like BERTScore and COMET. Among open-source models, Qwen2.5-VL-7B achieves the highest score (57.80), trailing only four commercial systems. Other models such as LLaVA-OneVision-7B (53.89) and InternVL2.5-8B (54.69) also perform competitively, with consistent gains observed as model size increases—for instance, Qwen2.5-VL-7B outperforms its 3B variant by 2.98, and InternVL2.5-8B surpasses 2B by 4.76. To isolate the effect of visual input, we compare LVLMs with their language-only counterparts. Visual grounding yields consistent improvements, e.g., +3.84 for Qwen2.5-VL-7B, +8.34 for InternVL2.5-8B, and +11.02 for InternLM2.5-1.8B. These trends hold across scales, confirming the utility of visual information in translation. Qwen models show strong multilingual ability even at smaller scales. In contrast, InternLM2.5 benefits more from visual context. These results highlight the essential role of visual grounding in effective multilingual vision-language translation.

\begin{table}[t]
    \centering
    \small 
    \setlength\tabcolsep{2pt} 
    \renewcommand{\arraystretch}{1.0}
    \resizebox{\linewidth}{!}{

    \begin{tabular}{l|ccc|ccc|ccc|c}
        \toprule
        \multicolumn{1}{c|}{\textbf{Method}} & \multicolumn{3}{c|}{\textbf{MIT-10M}} & \multicolumn{3}{c|}{\textbf{OCRMT30K}} & \multicolumn{3}{c|}{\textbf{MTIT6}} & \multicolumn{1}{c}{\textbf{Avg.}} \\
        \cmidrule{2-10} 
        & \textbf{BLEU} & \textbf{BS-F1} & \textbf{COMET} & \textbf{BLEU} & \textbf{BS-F1} & \textbf{COMET} & \textbf{BLEU} & \textbf{BS-F1} & \textbf{COMET} \\
        
        \midrule
        \multicolumn{10}{c}{\textbf{Closed-Source Models}} \\
        \midrule
        \rowcolor{lightblue}
        GPT-4o & 18.55 & 78.22	& 67.60	& 18.25	& 87.33	& 65.88	& 28.56	& \textbf{84.34}	& \textbf{80.65}	& 58.82 \\
        \rowcolor{lightblue}
        Gemini-2.0-flash & 17.08	& 77.99	& 65.27	& 16.62	& 86.76	& 62.15	& 23.63	& 82.73	& 77.92	& 56.68 \\
        \rowcolor{lightblue}
        Qwen-VL-Max & 21.72	& \textbf{80.21}	& \textbf{69.42}	& 17.30	& 87.29	& 65.58	& \textbf{30.07}	& 83.32	& 79.22	& \textbf{59.35} \\
        
        \rowcolor{lightblue}
        Claude3.7 & 8.22	& 74.61	& 61.84	& 10.49	& 73.52	& 60.16	& 9.72	& 77.45	& 70.38	& 49.60 \\
        \rowcolor{purple}
        GPT4-o1 & \textbf{21.81}	& 79.80	& 68.64	& 15.53	& 85.42	& 63.36	& 26.16	& 84.05	& 79.86	& 58.29 \\
        \rowcolor{purple}
        QvQ-72B-preview & 7.44	& 73.65	& 61.79	& 13.33	& 73.75	& 62.51	& 19.60	& 76.60	& 74.41	& 51.45 \\
        \rowcolor{lighterblue}
        Deepseek-v3 & 16.07	& 76.49	& 62.90	& \textbf{19.01}	& 88.34	& \textbf{66.14}	& 29.14	& 83.95	& 80.58	& 58.07 \\
        \rowcolor{lighterblue}
        Macro-MT & 13.01 & 73.56 & 59.58 & 16.95 & 86.61 & 64.28 & 20.70 & 80.46 & 76.92 & 54.67 \\
        \rowcolor{lighterblue}
        Qwen2.5-Max & 11.98	& 75.81	& 61.70	& 18.12	& \textbf{88.79}	& 65.80	& 24.86	& 83.30	& 79.59	& 56.66 \\
        \rowcolor{darkerblue}
        Deepseek-r1 & 11.95	& 74.85	& 61.65	& 17.57	& 86.87	& 66.05	& 19.69	& 82.46	& 79.56	& 55.63 \\
        \rowcolor{darkerblue}
        QwQ-32B & 10.00	& 73.89	& 60.53	& 10.23	& 86.03	& 65.29	& 18.37	& 81.82	& 78.34	& 53.83 \\
        \midrule
        \multicolumn{10}{c}{\textbf{Open-Source Models}} \\
        \midrule
        
        \rowcolor{lightblue}
        Qwen2.5-VL-7B & 17.84	& 77.24	& 64.15	& 17.78	& 88.62	& 65.52	& 25.53	& 83.39	& 80.09	& 57.80 \\
        \rowcolor{lightblue}
        \hspace*{1em}\footnotesize{\textit{-- w/o OCR}} & 20.74	& 78.27	& 66.53	& 15.77	& 87.13	& 63.24	& 21.94	& 81.45	& 77.99	& 57.01 \\
        
        \rowcolor{lightblue}
        LLaVA-OneVision-7B & 10.89	& 74.08	& 59.53	& 13.84	& 87.83	& 64.99	& 19.81	& 79.67	& 74.33	& 53.89 \\
        \rowcolor{lightblue}
        \hspace*{1em}\footnotesize{\textit{-- w/o OCR}} & 12.28	& 74.74	& 61.34	& 5.83	& 83.91	& 50.07	& 13.32	& 73.54	& 61.99	& 48.56 \\
        
        \rowcolor{lightblue}
        InternVL2.5-8B& 12.10	& 74.62	& 59.90	& 15.06	& 87.89	& 64.23	& 21.20	& 80.85	& 76.36	& 54.69 \\
        \rowcolor{lightblue}
        \hspace*{1em}\footnotesize{\textit{-- w/o OCR}} & 9.46	& 72.76	& 58.53	& 9.15	& 83.36	& 56.16	& 13.75	& 76.64	& 70.42	& 50.03 \\

        \rowcolor{lightblue}
        Qwen2.5-VL-3B & 13.95	& 74.65	& 61.02	& 13.36	& 86.46	& 61.67	& 21.51	& 80.29	& 75.60	& 54.28  \\
        \rowcolor{lightblue}
        \hspace*{1em}\footnotesize{\textit{-- w/o OCR}} & 14.42	& 74.48	& 61.78	& 9.44	& 82.21	& 56.71	& 17.59	& 78.34	& 74.30	& 52.14 \\

        \rowcolor{lightblue}
        InternVL2.5-2B & 9.09	& 72.17	& 56.40	& 8.51	& 83.93	& 57.56	& 14.71	& 77.07	& 69.92	& 49.93 \\
        \rowcolor{lightblue}
        \hspace*{1em}\footnotesize{\textit{-- w/o OCR}} & 5.61	& 69.99	& 53.58	& 6.05	& 81.26	& 52.71	& 12.86	& 74.36	& 67.50	& 47.10\\

        \rowcolor{lighterblue}
        Qwen2.5-7B-Instruct & 10.62	& 73.79	& 58.65	& 15.64	& 86.98	& 62.43	& 21.12	& 80.42	& 75.99	& 53.96\\
        \rowcolor{lighterblue}
        Qwen2-7B-Instruct& 6.66	& 71.88	& 55.58	& 10.86	& 87.37	& 63.61	& 18.43	& 80.33	& 75.59	& 52.26 \\
        \rowcolor{lighterblue}
        InternLM2.5-7B-Chat& 2.70	& 67.25	& 49.98	& 8.97	& 85.92	& 59.92	& 4.38	& 73.09	& 64.98	& 46.35\\
        \rowcolor{lighterblue}
        Qwen2.5-3B-Instruct  & 10.20	& 72.46	& 56.81	& 12.83	& 85.27	& 60.34	& 17.83	& 78.25	& 72.69	& 51.85 \\
        \rowcolor{lighterblue}
        InternLM2.5-1.8B-Chat  & 0.61	& 60.11	& 40.22	& 1.39	& 80.50	& 49.35	& 1.21	& 64.21	& 52.62	& 38.91\\
    
        \bottomrule
    \end{tabular}
    }
    \caption{Evaluation results of closed- \& open-source VLT models across MIT-10M, OCRMT30K, and MTIT6}
    \label{tab:eval_3dataset}
\end{table}

\subsection{OCR Dependency and Correction Ability}
End-to-end models leverage the built-in OCR capability of LVLMs to recognize and translate image text directly. In contrast, cascaded models rely on explicit OCR tools for text extraction. This raises a key question: Are current LVLMs sufficiently capable of OCR to eliminate the need for auxiliary tools?

To answer this, we evaluate three end-to-end models—Qwen2.5-VL (7B/3B), LLaVA-OneVision-7B, and InternVL2.5 (8B/2B)—under two conditions: (1) using only native OCR (w/o OCR), and (2) supplementing native OCR with tool-extracted text as reference (w/ OCR). Surprisingly, only Qwen2.5-VL-7B maintains strong performance without external OCR (57.01 w/o, 57.80 w/), and even shows minor gains in the more OCR-sensitive datasets (OCRMT30K, MTIT6) when provided with auxiliary input. All other models exhibit notable improvements under w/ OCR, with LLaVA-OneVision-7B underperforming even its own language-only counterpart when relying solely on internal OCR.

These findings indicate that, although end-to-end models are designed to operate without external OCR tools, current OCR capabilities in most LVLMs remain sub-optimal. Supplementing them with OCR-extracted text can still enhance performance. One exception is Qwen2.5-VL, which maintains strong results without auxiliary OCR input. We hypothesize that this robustness stems from its exposure to document-centric VQA tasks \cite{fu2024ocrbench, yang2024cc} and instruction-tuned reasoning over images during pretraining—enabling the model to directly infer textual content and context from raw visual input, rather than relying on intermediate OCR stages.

\subsection{Generation vs Reasoning Behavior}
Recent advances in reasoning-based models have significantly extended the capabilities of LLMs. Unlike generation-based models, which directly produce target translations based on input and prompts, reasoning models engage in a multi-step process: first interpreting input content (image or OCR text), then iteratively refining intermediate responses through task analysis, contextual reasoning, and validation before arriving at a final translation.

\begin{figure}[h]
  \centering
  \includegraphics[width=0.9\linewidth]{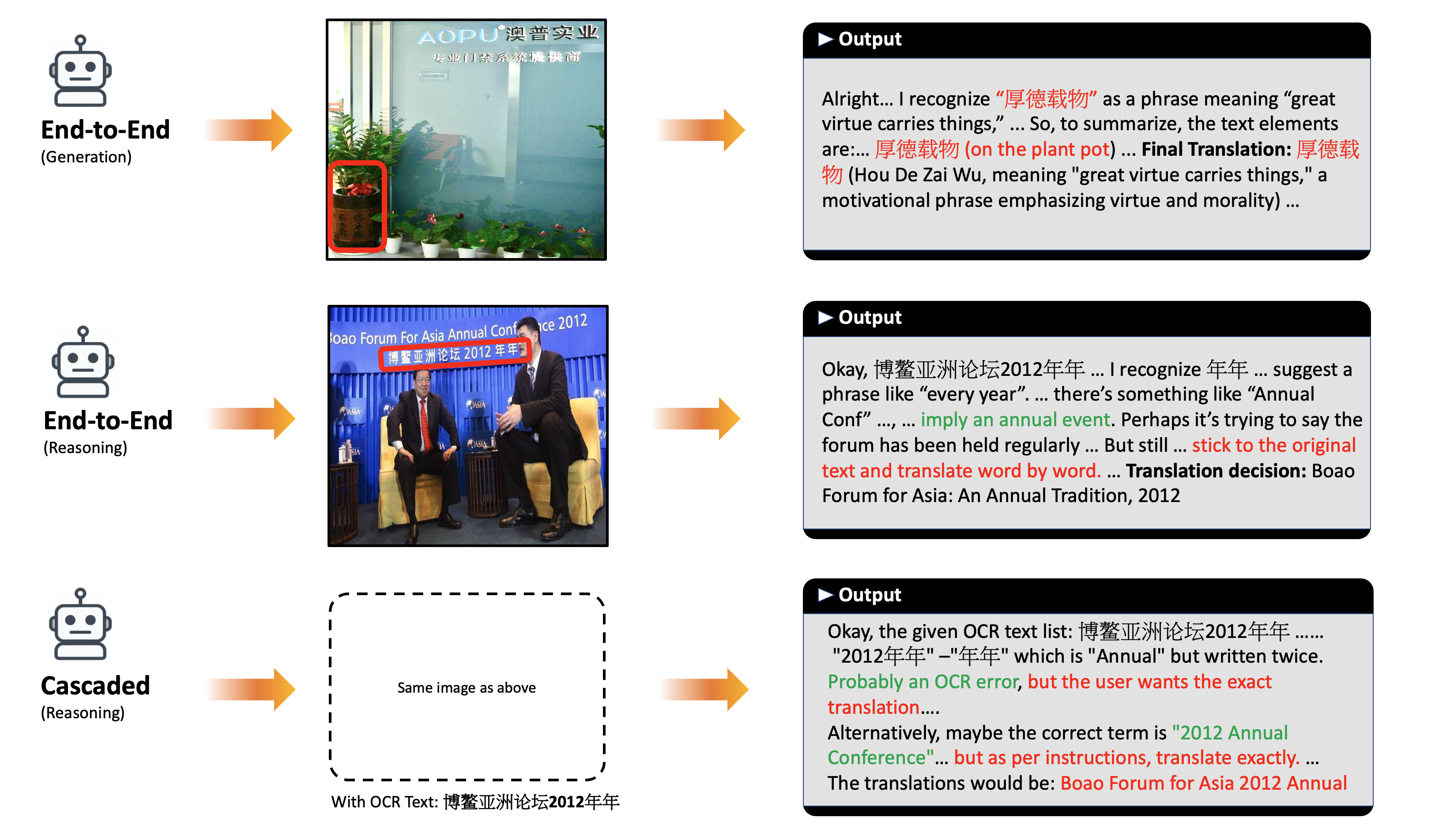}
  \caption{Behavioral Observations of Cascaded and End-to-End Models}
  \label{fig:behavior_example} 
\end{figure}

\begin{CJK*}{UTF8}{gbsn}
In our experiments, we observed substantial behavioral differences between the two model types. Generation models require strict and well-scoped prompts to function effectively \cite{hinck2024why}. These \textbf{strict-type} instructions must explicitly specify the task objective, constraints, and desired output format, as generation models produce one-shot outputs directly based on the instruction. In contrast, reasoning models tend to over-interpret such strict instructions. For instance, in Figure ~\ref{fig:behavior_example}, when translating an image containing partially occluded Chinese text “年会” (Annual Meeting), the character “会” was visually obstructed. A well-functioning reasoning model can infer the correct term based on partial visual cues and context. Similarly, in cascaded setups, the model might identify OCR errors such as “年年” (Year Year) and correct them by inferring the correct phrase. However, with strict prompts, the model may discard the correct inference due to over-adherence to the instruction, leading to reduced translation quality and longer inference time. We thus recommend using \textbf{collaborative-type} instructions for reasoning models—prompts that trust the model's reasoning capacity, encourage contextual inference, and avoid excessive constraint. This approach better aligns with their internal multi-stage deliberation process and leads to more accurate and fluent outputs.

Moreover, end-to-end models exhibit a unique vulnerability: \textbf{visual hallucinations}. These models, while capable of implicit OCR and grounding their translations in image content, can misinterpret ambiguous visual inputs. For example, when presented with an image of a typical Chinese office scene, a model hallucinated a decorative motto (“厚德载物”) on a flower pot that was visually indistinct. Such hallucinations can lead to factual errors in translation. To address this, we introduced anti-hallucination instructions during inference by simply adding a constraint: "\textit{Do not focus on visually unclear areas}." As shown in Table ~\ref{tab:instruction_types}, this significantly improved translation performance—Claude-3.7 (generation) improved by an average of 4.23, while QvQ-72B-Preview (reasoning) improved by 7.36 across three datasets.
\end{CJK*}

\begin{table}
\resizebox{\linewidth}{!}{

\begin{tabular}{llccc}
\toprule
\textbf{Model} & \textbf{Inst. Type} & \textbf{MIT-10M} & \textbf{OCRMT30K} & \textbf{MTIT6} \\
\midrule
\multirow{2}{*}{QvQ-72B-Preview}  
  & Strict  & 48.21 & 44.48 & 52.74 \\
  & Collabrative  & 51.39 & 49.86 & 56.87 \\
\midrule
\multirow{2}{*}{Claude-3.7} 
  & Strict  & 46.71 & 48.89 & 41.79 \\
  & Anti-Hal.  & 54.60 & 52.36 & 52.52 \\
\bottomrule
\end{tabular}
}
\caption{Impact of Strict, Collabrative and Anti-Hallucination Instruction Types on Model Performance Across Datasets}
\label{tab:instruction_types}
\end{table}

\section{Evaluation Challenges and Solutions}

\subsection{Metric Discrepancy in Context}
In Section 3.2, we evaluated model performance using commonly adopted metrics in VLT: BLEU, BERTScore, and COMET. These metrics capture different aspects of translation quality-BLEU emphasizes surface-level n-gram overlap, while BERTScore and COMET rely on contextual semantic similarity. However, a consistent and concerning discrepancy arises: semantic metrics often produce scores 3-8 times higher than BLEU, leading to a skewed arithmetic mean where BLEU is effectively underweighted. This imbalance is not incidental. The popularity of sentence-level metrics in VLT stems from the belief that lexical diversity and flexible phrasing in translation make strict n-gram matching less reliable. Nonetheless, we argue that context matters, especially in image-grounded scenarios. Consider the example where the source image contains the English word EXIT. A model that simply copies "EXIT" into the translation-failing to translate it into a target language-may still receive a COMET score above 0.8, due to cross-lingual semantic embeddings. However, BLEU correctly penalizes this failure. In short and phrase-level contexts, such as product labels, BLEU better reflects the precision required for accurate lexical choices. It captures failures that semantic metrics overlook. This leads to two fundamental questions: Which metrics truly reflect VLT performance under real-world constraints? If arithmetic averaging is misleading, how should multiple metrics be interpreted or combined?

    
    
    

\subsection{Metric-Human Alignment Across Density}

\begin{figure}[h]
  \centering
  \includegraphics[width=0.6\linewidth]{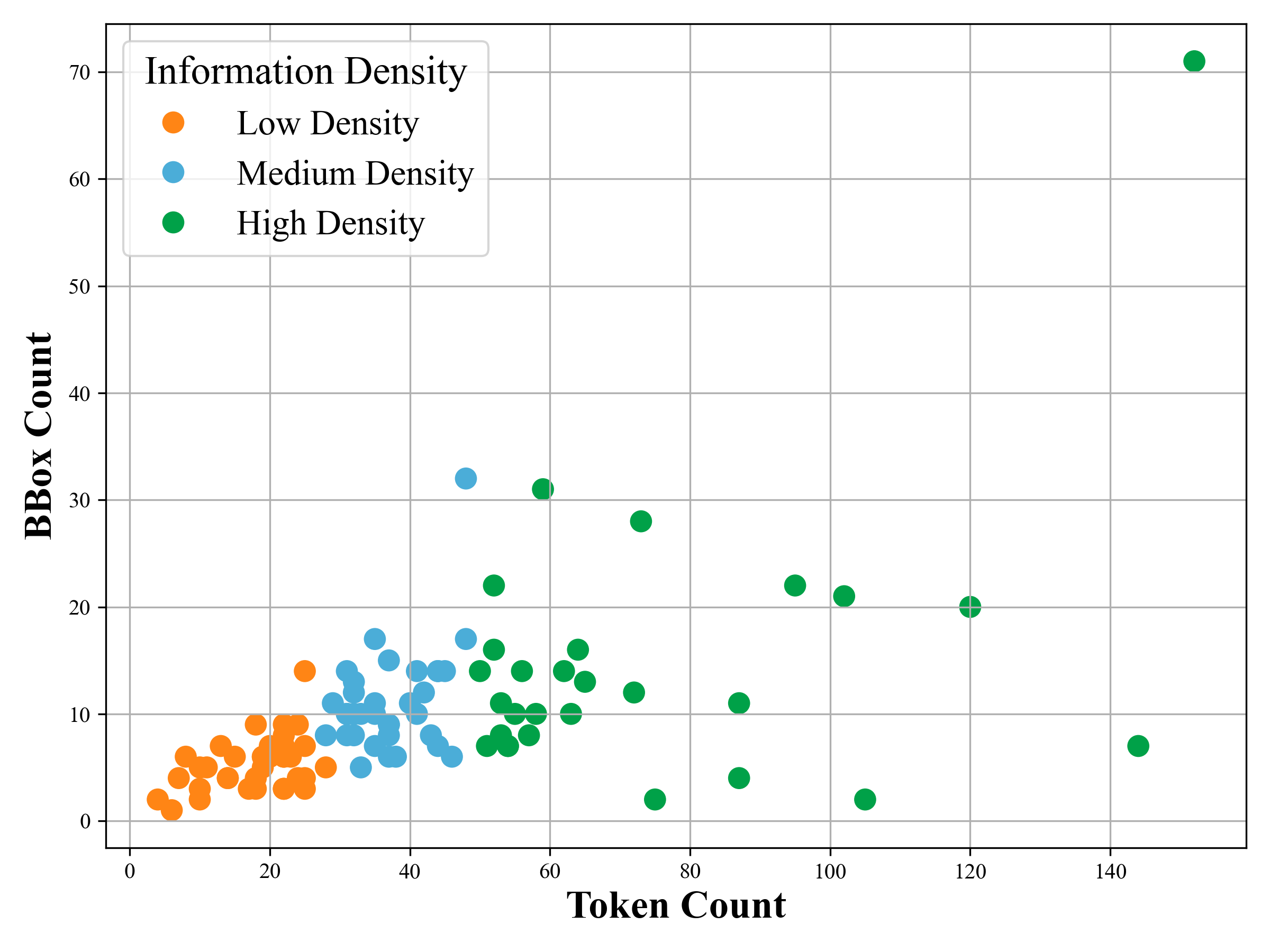}
  \caption{Information Density Clustering}
  \label{fig:info_density} 
\end{figure}

To systematically capture the variation, we propose an \textbf{Information Density} taxonomy based on the number of bounding boxes and the token length of the source sentence. Using K-means clustering over these two dimensions, we partition the samples into three categories: \textbf{Low Density}: $\leq 10$ bounding boxes, 1-30 tokens. \textbf{Medium Density}: 5-20 bounding boxes, 30-50 tokens. \textbf{High Density}: 8-30 bounding boxes, 50-90 tokens. This categorization reflects increasing complexity along both the visual and linguistic axes, as illustrated in Figure ~\ref{fig:info_density}.

To evaluate how well automatic metrics align with human judgment under different density levels, we collect manual ratings for translations generated by GPT-4o across all three datasets. Each translation is rated on a 1-5 scale across four dimensions-semantic adequacy, grammaticality, fluency, and cultural fit (as defined in Section 2.2)-with the overall score computed as their average. We first compute Pearson correlations between human ratings and several common metrics, including BLEU, CHRF, CHRF++, METEOR, TER, BERTScore, and COMET. Among them, BLEU, CHRF++, BERTScore, and COMET demonstrate the highest correlations (all > 0.4), indicating greater reliability for evaluating VLT. However, these metrics vary in performance across input density levels, prompting a more fine-grained analysis. To this end, we examine metric-human correlations across different levels of information density (Table ~\ref{tab:correlation}). BLEU performs best in low-density settings (0.47) but drops sharply in medium (0.26) and high (0.17), reflecting its sensitivity to n-gram overlap and limited robustness to structural variation. CHRF++ exhibits the opposite trend, improving in medium and high-density cases (both 0.41), likely due to its tolerance for lexical variation at the character level. BERTScore shows a non-monotonic pattern-strong in low (0.60) and high (0.44) densities, but unexpectedly weak in medium (0.26), possibly due to embedding instability in mid-length sentences. COMET maintains the highest and most consistent correlation overall, from 0.74 in low to 0.43 in high density, benefiting from its use of source, reference, and hypothesis, though longer sequences may introduce alignment challenges. These findings highlight that no single metric is uniformly reliable, and contextual complexity must be considered in evaluation.BLEU excels in simpler cases, while CHRF++ and COMET offer better robustness under increasing density. Embedding-based metrics such as BERTScore require careful interpretation, especially for mid-length samples. These results underscore the importance of density-aware evaluation for fair and accurate benchmarking of VLT systems

\begin{table}
\small
\resizebox{\linewidth}{!}{
    \begin{tabular}{lcccc}
    \toprule
    \textbf{Info. Density} & \textbf{BLEU} & \textbf{CHRF++} & \textbf{BS-F1} & \textbf{COMET} \\
    \midrule
    
    Low  & 0.4717 & 0.2965 & 0.5996 & 0.7360  \\
    
    Medium  & 0.2569 & 0.4130 & 0.2650 & 0.5822 \\
    
    High  & 0.1683 & 0.4131 & 0.4438 & 0.4269 \\
    \bottomrule
    \end{tabular}
    }
    \caption{Metric-Human Correlation by Information Density}
    \label{tab:correlation}
\end{table}

\subsection{Density-Aware Evaluation: Context Matters}

Previous analysis demonstrates that automatic metrics-BLEU, CHRF++, BERTScore, and COMET-exhibit varying reliability across different levels of information density. This observation underscores the need for \textit{granular, context-aware evaluation}, rather than relying on a simple arithmetic average across metrics. To address this, we propose the \textbf{Density-Aware Score (DA Score)}, a weighted combination of four evaluation metrics:
\begin{equation*}
\text{DA Score} = \alpha \cdot \text{BLEU} + \beta \cdot \text{CHRF++} + \lambda \cdot \text{BERT-F1} + \phi \cdot \text{COMET}.
\end{equation*}
Here, the weights $\alpha$, $\beta$, $\lambda$, and $\phi$ represent the \textit{relative importance} of each metric, derived from their \textit{correlation with human judgments} within each information density category (see Table~\ref{tab:correlation}). A higher weight indicates stronger alignment between the metric and human evaluation under that specific complexity level.

\begin{figure}[htbp]
  \centering
  \includegraphics[width=0.6\linewidth]{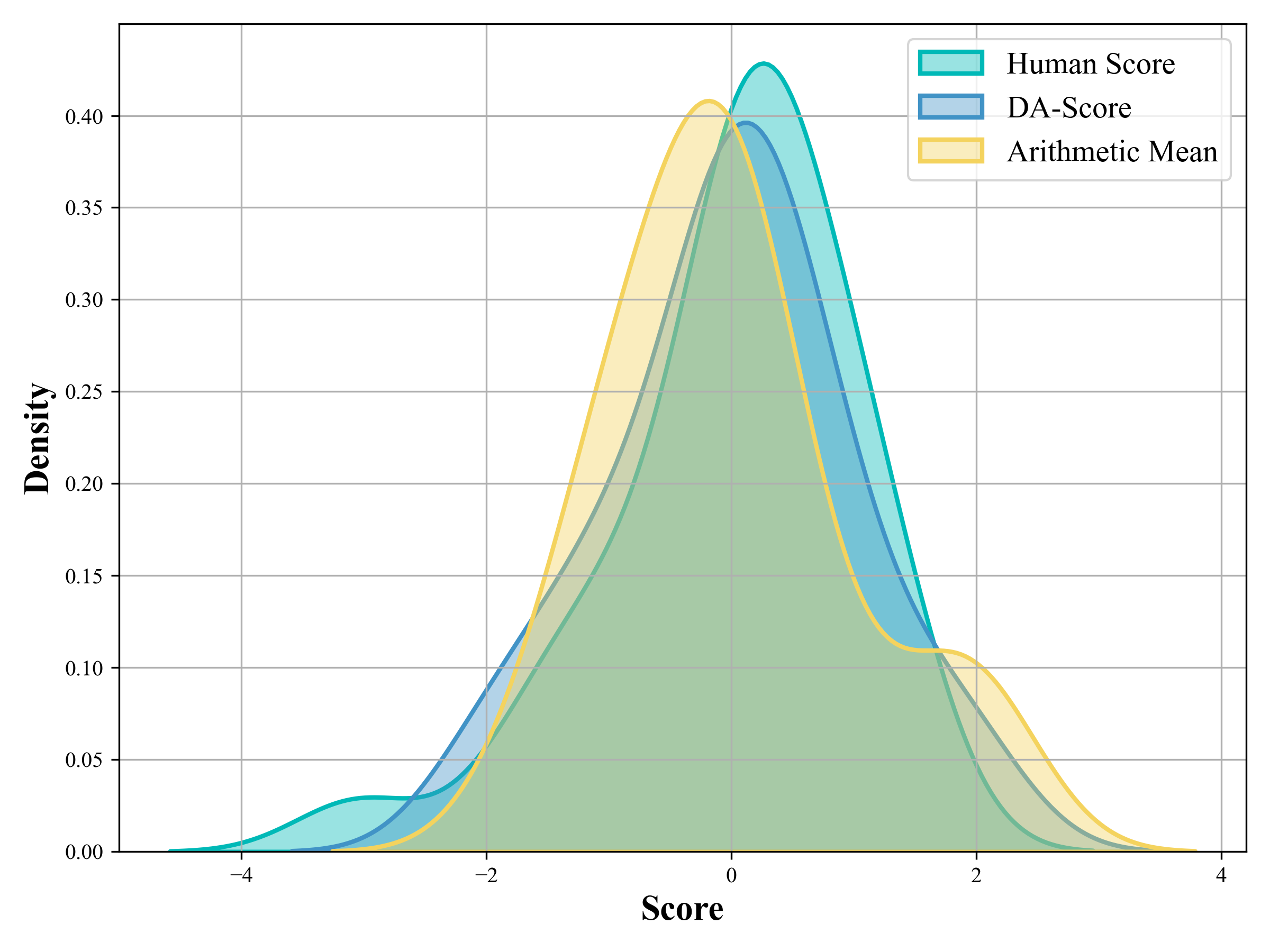}
  \caption{Score Distribution: Human vs DA-Score vs Arithmetic Mean}
  \label{fig:score_distribution} 
\end{figure}

As illustrated in Figure ~\ref{fig:score_distribution}, the kernel density plots compare the distribution of human evaluation scores (green), arithmetic mean scores (orange), and our proposed Density-Aware (DA) scores (blue). The DA score distribution shifts rightward compared to the arithmetic mean, aligning more closely with the human score curve. Notably, the human scores exhibit a longer left tail, reflecting a higher frequency of low-quality translations. In contrast, the arithmetic mean shows a shorter left tail, indicating an overly optimistic estimation that underrepresents lower-quality cases. The DA score, however, partially recovers this tail, suggesting that it better captures the quality variance observed in human judgments and corrects the optimistic bias introduced by naive averaging. By adapting metric weights to \textit{contextual difficulty}, the DA Score provides a more robust and comprehensive measure of VLT performance. 


\section{Benchmarking Multilingual VLT}

\subsection{Benchmark Framework Overview}

Our benchmark is built on AibTrans, a human-curated dataset containing 100 images and 6993 sentence-level translation instances, spanning seven target languages: English, German, Spanish, Russian, Arabic, Japanese, and Hindi-all translated from Chinese as the source language. AibTrans dataset exhibits diverse complexity, with the number of bounding boxes ranging from 1 to 38 and token lengths from 1 to 40. We evaluate both commercial and open-source models, covering end-to-end and cascaded architectures, which are the same as we used in Section 3.2. Following the analysis in Section 4, we report performance under low, medium, and high information density conditions using BLEU, CHRF++, BERTScore, COMET, and our proposed DA Score as a holistic measure of model capability.

\subsection{Fine-Tuning Strategies and Observations}
Fine-tuning open-source LVLMs can improve their alignment with the VLT task. However, an important question arises: should fine-tuning focus on improving task-specific adaptation using high-resource language pairs, or on increasing multilingual generalization? To explore this, we first conduct fine-tuning on the high-quality English-to-Chinese dataset OCRMT30K using the popular LoRA technique, applying it to Qwen2.5-VL (7B/3B) and InternVL2.5 (8B/2B). As shown in Table ~\ref{tab:ocrmt_sft}, fine-tuning on a high-resource language pair leads to performance degradation across both high- and low-resource translation directions. This trend holds across all model sizes. We attribute this to the fact that these LVLMs have already been extensively pre-trained and SFT-tuned on a wide range of multilingual and multimodal tasks. Fine-tuning on a single high-resource pair may harm the models' generalization and disrupt their learned multilingual representations.

\begin{table}[t]
    \centering
    \small 
    \setlength\tabcolsep{2pt} 
    \renewcommand{\arraystretch}{1.0}
    \resizebox{\linewidth}{!}{
    \begin{tabular}{l|ccc|ccc|ccc|c}
        \toprule
        \multicolumn{1}{c|}{\textbf{Method}} & \multicolumn{3}{c|}{\textbf{MIT-10M}} & \multicolumn{3}{c|}{\textbf{OCRMT30K}} & \multicolumn{3}{c|}{\textbf{MTIT6}} & \multicolumn{1}{c}{\textbf{Avg.}} \\
        \cmidrule{2-10} 
        & \textbf{BLEU} & \textbf{BS-F1} & \textbf{COMET} & \textbf{BLEU} & \textbf{BS-F1} & \textbf{COMET} & \textbf{BLEU} & \textbf{BS-F1} & \textbf{COMET} \\

        \midrule
        \multicolumn{10}{c}{\textbf{Open-Source Models Baselines}} \\
        \midrule
        \rowcolor{lightblue}

        Qwen2.5-VL-7B & 17.84	& 77.24	& 64.15	& 17.78	& 88.62	& 65.52	& 25.53	& 83.39	& 80.09	& 57.80 \\
        
        \rowcolor{lightblue}
        InternVL2.5-8B & 12.10	& 74.62	& 59.90	& 15.06	& 87.89	& 64.23	& 21.20	& 80.85	& 76.36	& 54.69 \\
   
        \rowcolor{lightblue}
        Qwen2.5-VL-3B & 13.95	& 74.65	& 61.02	& 13.36	& 86.46	& 61.67	& 21.51	& 80.29	& 75.60	& 54.28  \\
        
        \rowcolor{lightblue}
        InternVL2.5-2B & 9.09	& 72.17	& 56.40	& 8.51	& 83.93	& 57.56	& 14.71	& 77.07	& 69.92	& 49.93 \\
        
        \midrule
        \multicolumn{10}{c}{\textbf{SFT with LoRA OCRMT30K}} \\
        \midrule
        \rowcolor{lightblue}
        Qwen2.5-VL-7B-LoRA & 16.52	& 75.00	& 62.47	& 25.08	& 89.20	& 67.25	& 14.98	& 74.80	& 70.15	& 55.05 \\
    
        \rowcolor{lightblue}
        InternVL2.5-8B-LoRA & 8.37	& 71.10	& 54.82	& 19.42	& 88.80	& 64.83	& 6.29	& 68.40	& 59.29	& 49.04 \\
   
        \rowcolor{lightblue}
        Qwen2.5-VL-3B-LoRA & 12.37	& 73.00	& 59.55	& 18.22	& 87.60	& 63.11	& 9.44	& 72.80	& 65.21	& 51.26 \\

        \rowcolor{lightblue}
        InternVL2.5-2B-LoRA & 8.51	& 70.90	& 54.74	& 15.47	& 88.00	& 62.79	& 10.07	& 70.90	& 61.32	& 49.19 \\

        \bottomrule
    \end{tabular}
    }
    \caption{Performance of LVLMs fine-tuned on a high-resource language data with LoRA.}
    \label{tab:ocrmt_sft}
\end{table}

To address this, we adopt a \textbf{balanced multilingual fine-tuning} strategy, sampling 500 examples for each language pair from 8~$\rightarrow$~13 translation directions in MIT-10M. Results in Table ~\ref{tab:sample_sft} show that this strategy consistently improves performance across all models, especially those with smaller parameter sizes. For example, Qwen2.5-VL-3B gains +2.51 points in overall score after multilingual fine-tuning. We hypothesize that this is because VLT involves multiple sub-skills: OCR recognition via the visual encoder, multimodal fusion via adapters, and multilingual generation via the base LLM. Balanced multilingual exposure helps maintain multilingual competence while adapting to task-specific nuances.

We further examine the effect of training data volume. When increasing the sample size from 500 to 1000 per translation direction, model performance continues to improve-for instance, Qwen2.5-VL-7B shows a +3.46 point gain. However, further increasing the per-direction sample size to 2000 brings little additional benefit, suggesting diminishing returns. These findings indicate that a relatively small but balanced multilingual dataset is sufficient for adapting LVLMs to the VLT task, and excessive scaling may not be necessary. Fine-tuning should thus prioritize alignment with the VLT task structure rather than attempting to re-learn general multilingual competence.

\begin{table}[t]
    \centering
    \small 
    \setlength\tabcolsep{2pt} 
    \renewcommand{\arraystretch}{1.0}
    \resizebox{\linewidth}{!}{
    \begin{tabular}{l|ccc|ccc|ccc|c}
        \toprule
        \multicolumn{1}{c|}{\textbf{Method}} & \multicolumn{3}{c|}{\textbf{MIT-10M}} & \multicolumn{3}{c|}{\textbf{OCRMT30K}} & \multicolumn{3}{c|}{\textbf{MTIT6}} & \multicolumn{1}{c}{\textbf{Avg.}} \\
        \cmidrule{2-10} 
        & \textbf{BLEU} & \textbf{BS-F1} & \textbf{COMET} & \textbf{BLEU} & \textbf{BS-F1} & \textbf{COMET} & \textbf{BLEU} & \textbf{BS-F1} & \textbf{COMET} \\
        
        \midrule
        \multicolumn{10}{c}{\textbf{Sample 1000}} \\
        \midrule
        \rowcolor{lightblue}
        Qwen2.5-VL-7B & \textbf{26.94}	& \textbf{81.27}	& \textbf{71.18}	& \textbf{18.73}	& \textbf{88.24}	& \textbf{64.99}	& \textbf{26.98}	& \textbf{82.96}	& \textbf{79.58}	& \textbf{60.10} \\
        
        \rowcolor{lightblue}
        LLaVA-OneVision-7B & 15.53	& 77.46	& 63.87	& 13.51	& 87.32	& 62.95	& 18.95	& 79.44	& 72.72	& 54.64 \\
        
        \rowcolor{lightblue}
        InternVL2.5-8B & 14.85	& 76.59	& 62.01	& 16.64	& 88.57	& 64.92	& 22.84	& 81.65	& 76.46	& 56.06 \\
        
        \rowcolor{lightblue}
        Qwen2.5-VL-3B & 16.08	& 77.07	& 64.82	& 16.36	& 87.47	& 64.13	& 23.68	& 82.41	& 78.77	& 56.75 \\

        \rowcolor{lightblue}
        InternVL2.5-2B & 10.86	& 73.85	& 56.56	& 12.08	& 87.41	& 61.76	& 16.96	& 78.54	& 70.10	& 52.01 \\

        \midrule
        \multicolumn{10}{c}{\textbf{Sample 500}} \\
        \midrule
        \rowcolor{lightblue}
        Qwen2.5-VL-7B & 16.25	& 77.04	& 64.85	& 15.91	& 87.67	& 64.35	& 22.82	& 82.31	& 78.59	& 56.64 \\
        
        \rowcolor{lightblue}
        LLaVA-OneVision-7B & 15.48	& 77.34	& 63.96	& 13.49	& 87.28	& 63.01	& 17.23	& 78.38	& 70.52	& 54.08 \\
        
        \rowcolor{lightblue}
        InternVL2.5-8B & 14.07	& 76.17	& 61.36	& 16.56	& 88.58	& 65.03	& 23.34	& 81.52	& 76.60	& 55.91 \\
        
        \rowcolor{lightblue}
        Qwen2.5-VL-3B & 16.49	& 77.18	& 64.95	& 15.41	& 87.49	& 63.92	& 24.59	& 82.54	& 78.58	& 56.79 \\

        \rowcolor{lightblue}
        InternVL2.5-2B & 9.27	& 72.72	& 54.74	& 11.61	& 87.21	& 61.20	& 12.92	& 77.00	& 67.89	& 50.51 \\

        \bottomrule
    \end{tabular}
    }
    \caption{Performance of LVLMs with balanced multilingual fine-tuning using 500 vs. 1000 samples per language pair.}
    \label{tab:sample_sft}
\end{table}

\subsection{Unified Evaluation Protocol}
\begin{table*}[t]
    \centering
    \footnotesize 
    \setlength\tabcolsep{2.2pt} 
    \renewcommand{\arraystretch}{1.0}
    \begin{tabular}{l|ccccc|ccccc|ccccc}
        \toprule
        \multicolumn{1}{c|}{\textbf{Method}} & \multicolumn{5}{c|}{\textbf{High-Density}} & \multicolumn{5}{c|}{\textbf{Medium-Density}} & \multicolumn{5}{c}{\textbf{Low-Density}} \\
        \cmidrule{2-16} 
        & \textbf{BLEU} & \textbf{CHRF++} & \textbf{BS-F1} & \textbf{COMET} & \multicolumn{1}{c|}{\textbf{DA-Score}} & \textbf{BLEU} & \textbf{CHRF++} & \textbf{BS-F1} & \textbf{COMET} & \multicolumn{1}{c|}{\textbf{DA-Score}} & \textbf{BLEU} & \textbf{CHRF++} & \textbf{BS-F1} & \textbf{COMET} & \multicolumn{1}{c}{\textbf{DA-Score}} \\
        
        \midrule
        \multicolumn{16}{c}{\textbf{Closed-Source Models}} \\
        \midrule
        \rowcolor{lightblue}
        GPT-4o & 11.63 & 36.53 & 75.85 & 70.19 & 55.56 & 9.15 & 31.29 & 73.88 & 63.96 & 47.52 & 9.13 & 33.06 & 74.09 & 63.56 & 50.06 \\
        \rowcolor{lightblue}
        Gemini-2.0-flash & 18.22 & 45.38 & 80.15 & 72.32 & 60.78 & 12.91 & 40.71 & 77.65 & 66.82 & 52.47 & 11.11 & 37.89 & 76.29 & 62.33 & 51.38 \\
        \rowcolor{lightblue}
        Qwen-VL-Max & 15.89 & 43.29 & 79.65 & 73.73 & 60.17 & 12.81 & 38.71 & 78.03 & 68.88 & 52.77 & 9.70 & 34.66 & 75.87 & 65.85 & 51.72 \\
        \rowcolor{lightblue}
        Claude3.7 & 11.71 & 31.01 & 74.97 & 66.56 & 52.66 & 10.21 & 29.25 & 73.79 & 63.44 & 46.93 & 8.37 & 27.01 & 73.22 & 61.83 & 48.18 \\
        \rowcolor{purple}
        GPT4-o1 & 17.62 & 44.92 & 79.35 & 73.63 & 60.72 & 14.34 & 41.90 & 78.62 & 70.26 & 54.53 & 12.87 & 40.25 & 78.03 & 67.85 & 54.53 \\
        \rowcolor{purple}
        QvQ-72B-preview & 9.98 & 33.79 & 73.84 & 66.00 & 52.74 & 8.98 & 33.33 & 73.25 & 64.28 & 48.06 & 8.80 & 33.37 & 74.60 & 64.69 & 50.57 \\
        \rowcolor{lighterblue}
        Deepseek-v3 & \textbf{25.24} & \textbf{50.29} & \textbf{82.65} & 75.93 & \textbf{64.82} & \textbf{19.77} & \textbf{45.37} & \textbf{80.32} & 69.66 & \textbf{56.46} & \textbf{16.92} & \textbf{43.06} & \textbf{79.42} & \textbf{67.42} & \textbf{56.08} \\
        \rowcolor{lighterblue}
        Macro-MT & 5.90 & 20.08 & 70.89 & 57.22 & 44.88 & 5.16 & 18.40 & 70.03 & 51.90 & 38.03 & 5.25 & 19.85 & 71.11 & 56.21 & 43.91 \\
        \rowcolor{lighterblue}
        Qwen2.5-Max & 20.58 & 47.60 & 81.09 & 74.49 & 62.61 & 15.33 & 42.13 & 78.81 & 69.00 & 54.31 & 12.57 & 39.38 & 77.72 & 65.41 & 53.40 \\
        \rowcolor{darkerblue}
        Deepseek-r1 & 19.14 & 45.71 & 81.14 & \textbf{76.01} & 62.37 & 14.34 & 41.79 & 78.70 & \textbf{70.43} & 54.58 & 14.21 & 41.31 & 78.62 & \textbf{68.01} & 55.21 \\
        \rowcolor{darkerblue}
        QwQ-32B & 15.73 & 42.95 & 79.79 & 74.60 & 60.36 & 11.66 & 38.33 & 77.18 & 69.02 & 52.38 & 10.84 & 36.54 & 76.89 & 66.49 & 52.76 \\
        \midrule
        \multicolumn{16}{c}{\textbf{Open-Source Models}} \\
        \midrule

        \rowcolor{lightblue}
        Qwen2.5-VL-7B & 9.21 & 29.10 & 74.78 & 63.31 & 50.81 & 9.05 & 30.06 & 74.63 & 62.01 & 46.55 & 8.49 & 29.22 & 74.21 & 61.26 & 48.60 \\
        
        \rowcolor{lightblue}
        LLaVA-OneVision-7B & 8.51 & 26.18 & 70.78 & 57.63 & 47.01 & 6.99 & 24.11 & 70.05 & 53.00 & 40.32 & 5.82 & 21.67 & 69.92 & 53.65 & 43.06 \\

        \rowcolor{lightblue}
        InternVL2.5-8B & 9.80 & 30.39 & 74.38 & 60.38 & 50.26 & 7.54 & 27.25 & 72.83 & 56.60 & 43.14 & 6.08 & 24.73 & 72.13 & 54.72 & 44.55 \\
        
        \rowcolor{lightblue}
        Qwen2.5-VL-3B & 4.84 & 19.40 & 70.31 & 55.74 & 43.96 & 4.86 & 19.79 & 70.24 & 54.20 & 39.28 & 4.61 & 20.52 & 70.17 & 55.10 & 43.20 \\

        \rowcolor{lightblue}
        InternVL2.5-2B & 4.24 & 16.27 & 67.47 & 48.49 & 40.00 & 2.92 & 13.13 & 66.47 & 45.73 & 33.23 & 1.81 & 10.54 & 65.68 & 45.21 & 36.43 \\

        \rowcolor{lighterblue}
        Qwen2-7B-Instruct & 12.97 & 37.75 & 77.36 & 68.17 & 55.93 & 9.08 & 32.55 & 74.83 & 63.10 & 47.68 & 6.97 & 29.02 & 73.84 & 59.92 & 47.66 \\
        \rowcolor{lighterblue}
        InternLM2.5-7B-Chat & 7.26 & 27.89 & 71.88 & 56.38 & 47.32 & 5.21 & 24.17 & 69.64 & 52.34 & 39.71 & 4.15 & 21.86 & 69.07 & 50.52 & 41.37 \\
        \rowcolor{lighterblue}
        Qwen2.5-3B-Instruct & 5.97 & 22.26 & 71.34 & 57.00 & 45.59 & 6.03 & 23.46 & 71.39 & 55.55 & 41.20 & 5.33 & 23.48 & 71.79 & 55.07 & 44.23 \\
        \rowcolor{lighterblue}
        InternLM2.5-1.8B-Chat & 2.07 & 15.72 & 64.62 & 44.20 & 37.46 & 1.56 & 13.45 & 62.39 & 41.33 & 30.69 & 1.18 & 11.80 & 62.71 & 41.32 & 34.26 \\
        
        \midrule
        \multicolumn{16}{c}{\textbf{SFT with LoRA}} \\
        \midrule

        \rowcolor{lightblue}
        Qwen2.5-VL-7B-LoRA & 8.67 & 28.58 & 74.18 & 65.63 & 51.10 & 8.72 & 29.59 & 74.54 & 64.56 & 47.33 & 9.12 & 31.01 & 75.18 & 65.50 & 50.75 \\

        \rowcolor{lightblue}
        LLaVA-OneVision-7B-LoRA & 9.32 & 29.58 & 73.91 & 60.08 & 49.75 & 6.51 & 25.61 & 72.16 & 55.46 & 41.96 & 5.97 & 24.81 & 72.34 & 55.62 & 44.91 \\

        \rowcolor{lightblue}
        InternVL2.5-8B-LoRA & 11.81 & 34.91 & 76.69 & 63.37 & 53.37 & 9.48 & 32.62 & 75.48 & 60.19 & 46.77 & 7.95 & 30.79 & 75.04 & 58.86 & 48.10 \\

        \rowcolor{lightblue}
        Qwen2.5-VL-3B-LoRA & 8.58 & 27.48 & 73.67 & 62.35 & 49.66 & 8.12 & 28.46 & 73.94 & 60.91 & 45.41 & 7.86 & 28.09 & 73.87 & 60.91 & 48.08 \\
        
        \rowcolor{lightblue}
        InternVL2.5-2B-LoRA & 6.46 & 26.73 & 72.07 & 51.10 & 45.40 & 5.53 & 24.39 & 71.39 & 49.98 & 39.22 & 4.23 & 22.15 & 71.23 & 50.06 & 41.88 \\

        \bottomrule
    \end{tabular}
    \caption{Unified Evaluation Protocol: DA-Score and other metrics results of different models evaluated under three levels of information density.}
    \label{tab:da_score_all}
\end{table*}

Table ~\ref{tab:da_score_all} summarizes the unified evaluation protocol across high, medium, and low information density tiers, revealing several important insights. First, the overall model performance aligns with earlier observations (Section 3.2). Among closed-source models, the cascaded model Deepseek-v3 consistently outperforms other commercial models across all density levels. End-to-end models such as Qwen-VL-Max and GPT4-o1 also deliver strong results, particularly in high-density scenarios where richer visual and contextual information supports more accurate translation. Open-source LVLMs, especially Qwen2.5-VL-7B, show promising results, indicating their potential to approach commercial-level performance even without task-specific tuning. Second, density-aware evaluation unveils finer-grained performance patterns that are obscured by traditional averaging. In particular, we observe an unexpected dip in performance on medium-density samples-often lower than that on low-density ones-across both open- and closed-source models. This challenges the assumption that shorter inputs are easier to translate. In low-density cases, although text spans are short, the lack of contextual cues places greater demands on the model to produce precise and semantically grounded translations. High-density inputs, on the other hand, provide ample linguistic and visual context that helps guide the translation. Medium-density samples sit in an ambiguous zone: they lack sufficient context for disambiguation, while still requiring precise handling of localized, visually grounded text. This complexity leads to inconsistent performance and highlights the importance of context-aware evaluation. Our DA Score framework proves effective in capturing such variations, offering a more faithful reflection of model behavior under different information conditions.

Finally, we find that fine-tuning brings consistent gains-particularly for smaller models. For example, Qwen2.5-VL-3B improves from 43.96 to 49.66 in high-density DA Score, while InternVL2.5-2B improves from 40.00 to 45.40. These results reaffirm the effectiveness of our balanced multilingual fine-tuning strategy: even with just 1000 samples per language direction, models significantly improve their ability to handle diverse translation scenarios without compromising their multilingual capacity.

\section{Related Work}

\textbf{Multilingual LVLMs and LLMs.} Early LVLMs \cite{liu2023visual, dai2023instructblip, zhu2023minigpt, wang-etal-2024-mitigating} were predominantly monolingual, often focusing on English. Recent progress in multilingual LLMs-including Qwen2.5-Instruct, InternLM2.5-Chat, and Deepseek-v3-has significantly broadened the multilingual capacity of LVLMs. Corresponding LVLMs such as Qwen2.5-VL, InternVL2.5, and Deepseek-VL2 now demonstrate strong performance across a wider range of languages. Commercial systems have also begun to exhibit advanced multilingual and multimodal capabilities. Models like GPT-4o, Claude3.7, and QwQ-32B showcase impressive performance in both multilingual generation and multimodal reasoning. In parallel, LVLMs have improved their multilingual OCR capabilities. Although the training details of these models are not publicly available, technical reports suggest that they place considerable emphasis on multilingual OCR supervision. For example, InternVL2.5 is reported to use 34 OCR-related datasets covering diverse languages, while LLaVA-OneVision employs 23 such datasets. However, systematic evaluation of these multilingual LVLMs on VLT tasks-especially their performance across diverse language directions-remains limited. Our work fills this gap by providing a comprehensive analysis of both commercial and open-source systems on the multilingual VLT task, revealing their strengths, limitations, and implications for future development.

\textbf{Vision-Language Translation.} VLT also known as text-in-image translation \cite{salesky2024benchmarking}, is a particularly challenging multimodal task that requires recognizing and translating text embedded within images. Prior work \cite{niu-etal-2024-umtit, lan2024translatotron, liang-etal-2024-document} has explored both end-to-end and cascaded approaches, each with distinct trade-offs. Early efforts often favored end-to-end designs-for example, \citet{lan2023exploring} employed a multimodal codebook to distinguish text regions from surrounding visual content, while AnyTrans \cite{qian2024anytrans} leveraged external OCR tools alongside visual cues to enhance translation accuracy. These works generally operate under the assumption that cascaded architectures are prone to error propagation between the OCR and translation stages. However, our findings challenge this assumption. We observe that cascaded systems, when equipped with strong multilingual LLMs, perform competitively-often benefiting from the LLM's ability to correct OCR errors via contextual reasoning. Furthermore, existing VLT studies rely on earlier-generation LVLMs and seldom evaluate modern multilingual LVLMs or multilingual LLMs. Our work fills this gap by systematically benchmarking state-of-the-art multilingual LVLMs and LLMs on the VLT task.


\section{Conclusion}

In this study, we present a comprehensive investigation of the VLT task, addressing key limitations in datasets, architectures, and evaluation metrics. We introduce AibTrans, a multilingual, parallel dataset with corrected OCR and culturally aligned translations. Through comparisons of end-to-end and cascaded models, we reveal OCR dependencies and distinct behaviors between generation and reasoning paradigms. To address evaluation challenges, we propose Density-Aware Evaluation for context-sensitive scoring. Finally, we present a novel VLT benchmark that incorporates high-quality data, density-aware scoring, and a balanced multilingual fine-tuning strategy for effective LVLM adaptation.

\bibliographystyle{ACM-Reference-Format}
\bibliography{sample-base}

\appendix


\section{The Masking Effect of Averaging}
To further investigate cross-lingual dynamics in VLT, we categorize the target languages in AibTrans into three resource tiers (following \cite{joshi2020state}): high-resource (English), medium-resource (German, Spanish, Japanese), and low-resource (Arabic, Russian, Hindi). As shown in Table ~\ref{tab:language_quality}, both end-to-end and cascaded models exhibit significantly better performance on high-resource languages compared to medium- and low-resource ones. In most models, the performance trend follows an intuitive hierarchy: \textbf{high} $>$ \textbf{medium} $>$ \textbf{low}. This is consistent across evaluation metrics and architectures. These results reveal a critical caveat in evaluation practice: averaging across languages can obscure important cross-lingual disparities. In particular, strong performance on high-resource languages may mask underperformance on low-resource targets, leading to over-optimistic conclusions about model capability. This masking effect highlights the need for disaggregated, language-level reporting in multilingual VLT evaluation-especially when assessing progress on low-resource and culturally distant languages.

\begin{table}[t]  
    \centering
    \small
    \setlength\tabcolsep{4pt} 
    \renewcommand{\arraystretch}{1.0}
    \resizebox{\linewidth}{!}{

    \begin{tabular}{l|>{\centering\arraybackslash}p{1.8cm}
                     >{\centering\arraybackslash}p{1.8cm}
                     >{\centering\arraybackslash}p{1.8cm}}
        \toprule
        \multicolumn{1}{c|}{\textbf{Method}} & \multicolumn{3}{c}{\textbf{Resources Quality of Languages}} \\
        \cmidrule{2-4} 
        & \textbf{High} & \textbf{Medium} & \textbf{Low} \\
        \midrule
        \rowcolor{lightblue}
        GPT-4o            & 57.05 & 49.47 & 49.37 \\
        \rowcolor{lightblue}
        Gemini-2.0-flash  & 56.57 & 53.92 & 51.86 \\
        \rowcolor{lightblue}
        Qwen-VL-Max       & 58.13 & 54.17 & 51.97 \\
        \rowcolor{lightblue}
        Claude3.7         & 57.79 & 50.17 & 45.73 \\
        \rowcolor{purple}
        GPT4-o1                & 59.13 & 54.94 & 54.28 \\
        \rowcolor{purple}
        QvQ-72B-preview   & 54.02 & 49.99 & 48.62 \\
        \rowcolor{lighterblue}
        Deepseek-v3       & 64.34 & 56.23 & 58.24 \\
        \rowcolor{lighterblue}
        Macro-MT          & 51.00 & 43.31 & 41.49 \\
        \rowcolor{lighterblue}
        Qwen2.5-Max       & 61.46 & 55.04 & 53.95 \\
        \rowcolor{darkerblue}
        Deepseek-r1       & 62.94 & 54.59 & 55.53 \\
        \rowcolor{darkerblue}
        QwQ-Plus          & 59.89 & 54.21 & 51.95 \\
        \bottomrule
    \end{tabular}
    }
    \caption{VLT performance across high-, medium-, and low-resource languages}
    \label{tab:language_quality}
\end{table}

\section{Trade-offs of Performance, Cost and Time}

\begin{figure}[h]
  \centering
  \includegraphics[width=\linewidth]{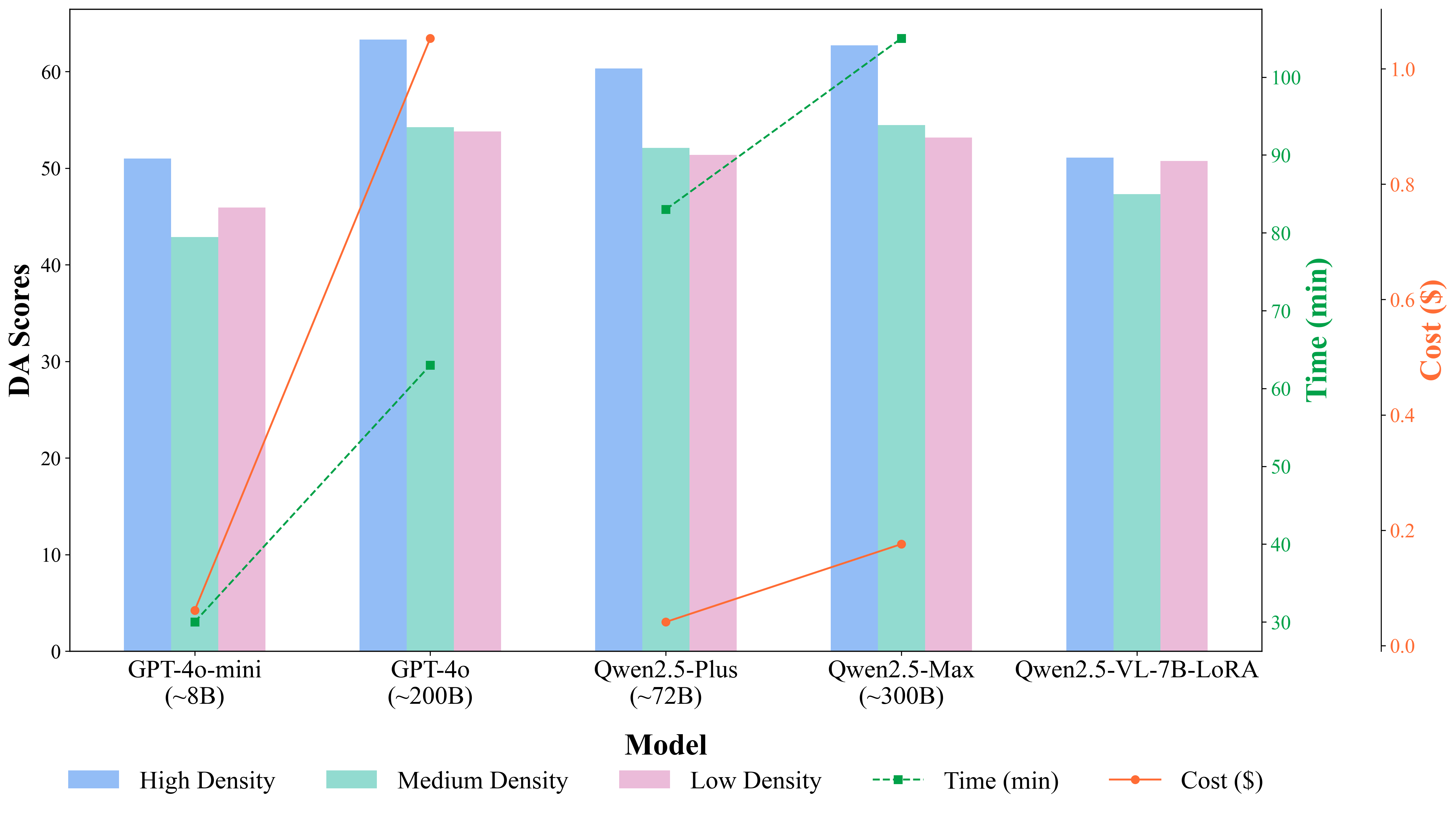}
  \caption{Closed-Source Model Trade-offs: DA Scores vs Cost \& Time}
  \label{fig:cost_eval2} 
\end{figure}

To assess the trade-offs between performance, inference time, and cost when using closed-source models for the VLT task, we evaluate two model families: OpenAI's GPT-4o ($\sim$200B) and GPT-4o-mini ($\sim$8B), and Alibaba's Qwen2.5-Max ($\sim$300B) and Qwen2.5-Plus ($\sim$72B). Each pair includes a larger and a smaller variant within the same architecture, enabling a comparison of scale effects.

As illustrated in Figure ~\ref{fig:cost_eval2}, GPT-4o and Qwen2.5-Max achieve the highest Density-Aware (DA) scores across all information density levels (high, medium, and low), indicating superior translation quality. Despite being smaller, Qwen2.5-Plus delivers competitive performance. Notably, the fine-tuned open-source model Qwen2.5-VL-7B-LoRA matches GPT-4o-mini's performance on high-density and outperforms it on medium- and low-density, demonstrating that well-tuned open-source models can also perform competitively.

While larger models generally yield better performance, the associated resource costs scale disproportionately. For instance, upgrading from GPT-4o-mini to GPT-4o results in a 24\% increase in DA score but incurs approximately 17$\times$ higher cost and more than 2$\times$ longer inference time. Similarly, moving from Qwen2.5-Plus to Qwen2.5-Max yields only a 4\% performance improvement, while inference time increases by about 27\% and cost becomes over four times higher. These findings suggest that mid-sized models like Qwen2.5-Plus offer a more practical trade-off between performance, cost, and inference time.

\section{Attribute Distribution of AibTrans}
\begin{figure}[h]
  \centering
  \includegraphics[width=\linewidth]{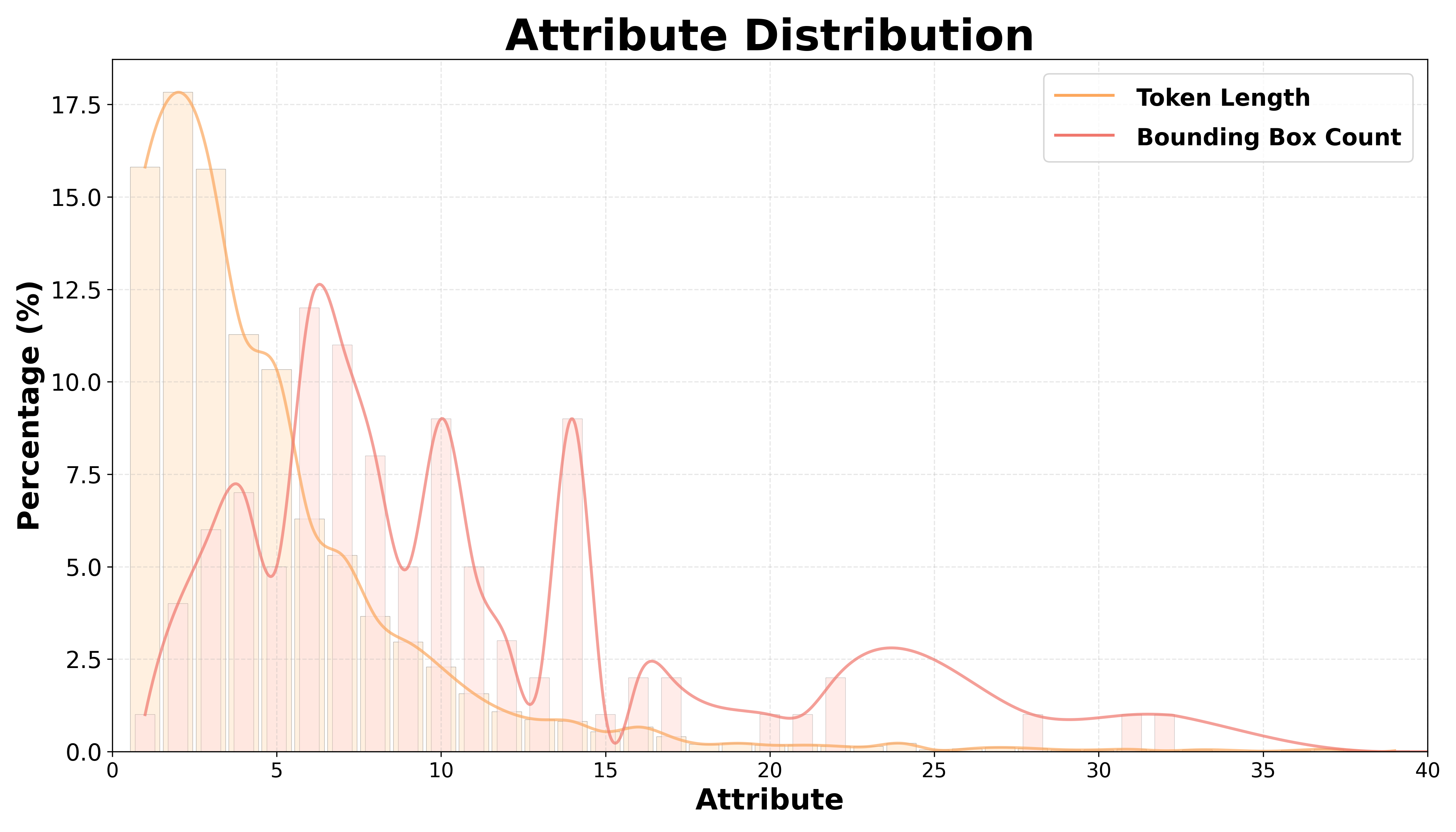}
  \caption{Attribute Distribution of AibTrans Dataset}
  \label{fig:attribute_distribution} 
\end{figure}

\section{Comparison of Existing Datasets}

\begin{table*}[ht]
\centering
\small
\renewcommand{\arraystretch}{1.2}
\setlength{\tabcolsep}{6pt}

\small

\begin{tabular}{
    l
    c
    c
>{\centering\arraybackslash}p{2cm}
    c
    c
    c
>{\centering\arraybackslash}p{2cm}
>{\centering\arraybackslash}p{2cm}
    c
}
\toprule
\textbf{Dataset} & 
\textbf{Train} & 
\textbf{Eval} & 
\textbf{OCR Tool} & 

\shortstack{\textbf{OCR} \\\textbf{Corrected}} & 
\textbf{Translation} & 
\shortstack{\textbf{Image}\\\textbf{Grounded}}&
\shortstack{\textbf{Source}\\\textbf{Languages}} & 
\shortstack{\textbf{Target}\\\textbf{Languages}} & 
\textbf{Parallel} \\
\midrule
\textbf{MIT-10M} & 
10M & 
10.4K & 
EasyOCR + GPT-4 & 
\xmark & 
Machine & 
\xmark & 
en, zh, ru, ar, hi, ko, th, tr & 
en, zh, es, fr, de, ru, ar, hi, ko, ja, it, th, tr, pt & 
\cmark \\

\textbf{OCRMT30K} & 
30K & 
1.2K & 
PaddleOCR & 
\xmark & 
Human & 
\cmark & 
zh & 
en & 
\xmark \\

\textbf{MTIT6} & 
--- & 
6K & 
PaddleOCR & 
\xmark & 
Human & 
\cmark & 
en, zh, ko, ja & 
en, zh, ko, ja & 
\xmark \\

\textbf{AibTrans} & 
--- & 
700/7000 & 
PaddleOCR + Hunman & 
\cmark & 
Human & 
\cmark & 
zh & 
en, de, ar, ja, ru, es, hi & 
\cmark \\

\bottomrule
\end{tabular}
\caption{Comparison of existing datasets for Vision-Language Translation.}

\label{tab:dataset_summary}
\end{table*}

\section{Extended experimental results of different models with LoRA}

\begin{table*}[t]
    \centering
    \footnotesize 
    \setlength\tabcolsep{2pt} 
    \renewcommand{\arraystretch}{0.95} 
    \resizebox{0.7\textwidth}{!}{ 
    \begin{tabular}{l|ccc|ccc|ccc|c}
        \toprule
        \multicolumn{1}{c|}{\textbf{Method}} & \multicolumn{3}{c|}{\textbf{MIT-10M}} & \multicolumn{3}{c|}{\textbf{OCRMT30K}} & \multicolumn{3}{c|}{\textbf{MTIT6}} & \multicolumn{1}{c}{\textbf{Avg.}} \\
        \cmidrule{2-10} 
        & \textbf{BLEU} & \textbf{BS-F1} & \textbf{COMET} & \textbf{BLEU} & \textbf{BS-F1} & \textbf{COMET} & \textbf{BLEU} & \textbf{BS-F1} & \textbf{COMET} \\
        
        \midrule
        \multicolumn{10}{c}{\textbf{Close-Source Models}} \\
        \midrule
        \rowcolor{lightblue}
        GPT-4o & 18.55 & 78.22  & 67.60 & 18.25 & 87.33 & 65.88 & 28.56 & \textbf{84.34}  & \textbf{80.65}  & 58.82 \\
        \rowcolor{lightblue}
        Gemini-2.0-flash & 17.08  & 77.99 & 65.27 & 16.62 & 86.76 & 62.15 & 23.63 & 82.73 & 77.92 & 56.68 \\
        \rowcolor{lightblue}
        Qwen-VL-Max & 21.72 & 80.21 & 69.42 & 17.30 & 87.29 & 65.58 & \textbf{30.07}  & 83.32 & 79.22 & \textbf{59.35} \\
        
        \rowcolor{lightblue}
        Claude3.7 & 8.22  & 74.61 & 61.84 & 10.49 & 73.52 & 60.16 & 9.72  & 77.45 & 70.38 & 49.60 \\
        \rowcolor{purple}
        GPT4-o1 & 21.81 & 79.80 & 68.64 & 15.53 & 85.42 & 63.36 & 26.16 & 84.05 & 79.86 & 58.29 \\
        \rowcolor{purple}
        QvQ-72B-preview & 7.44  & 73.65 & 61.79 & 13.33 & 73.75 & 62.51 & 19.60 & 76.60 & 74.41 & 51.45 \\
        \rowcolor{lighterblue}
        Deepseek-v3 & 16.07 & 76.49 & 62.90 & \textbf{19.01}  & 88.34 & \textbf{66.14}  & 29.14 & 83.95 & 80.58 & 58.07 \\
        \rowcolor{lighterblue}
        Macro-MT & 13.01 & 73.56 & 59.58 & 16.95 & 86.61 & 64.28 & 20.70 & 80.46 & 76.92 & 54.67 \\
        \rowcolor{lighterblue}
        Qwen2.5-Max & 11.98 & 75.81 & 61.70 & 18.12 & \textbf{88.79}  & 65.80 & 24.86 & 83.30 & 79.59 & 56.66 \\
        \rowcolor{darkerblue}
        Deepseek-r1 & 11.95 & 74.85 & 61.65 & 17.57 & 86.87 & 66.05 & 19.69 & 82.46 & 79.56 & 55.63 \\
        \rowcolor{darkerblue}
        QwQ-32B & 10.00 & 73.89 & 60.53 & 10.23 & 86.03 & 65.29 & 18.37 & 81.82 & 78.34 & 53.83 \\
        \midrule
        \multicolumn{10}{c}{\textbf{Open-Source Models}} \\
        \midrule
        
        \rowcolor{lightblue}
        Qwen2.5-VL-7B & 17.84 & 77.24 & 64.15 & 17.78 & 88.62 & 65.52 & 25.53 & 83.39 & 80.09 & 57.80 \\
        \rowcolor{lightblue}
        \hspace*{1em}\footnotesize{\textit{-- w/o OCR}} & 20.74 & 78.27 & 66.53 & 15.77 & 87.13 & 63.24 & 21.94 & 81.45 & 77.99 & 57.01 \\
        
        \rowcolor{lightblue}
        LLaVA-OneVision-7B & 10.89  & 74.08 & 59.53 & 13.84 & 87.83 & 64.99 & 19.81 & 79.67 & 74.33 & 53.89 \\
        \rowcolor{lightblue}
        \hspace*{1em}\footnotesize{\textit{-- w/o OCR}} & 12.28 & 74.74 & 61.34 & 5.83  & 83.91 & 50.07 & 13.32 & 73.54 & 61.99 & 48.56 \\
        
        \rowcolor{lightblue}
        InternVL2.5-8B& 12.10 & 74.62 & 59.90 & 15.06 & 87.89 & 64.23 & 21.20 & 80.85 & 76.36 & 54.69 \\
        \rowcolor{lightblue}
        \hspace*{1em}\footnotesize{\textit{-- w/o OCR}} & 9.46  & 72.76 & 58.53 & 9.15  & 83.36 & 56.16 & 13.75 & 76.64 & 70.42 & 50.03 \\

        \rowcolor{lightblue}
        Qwen2.5-VL-3B & 13.95 & 74.65 & 61.02 & 13.36 & 86.46 & 61.67 & 21.51 & 80.29 & 75.60 & 54.28  \\
        \rowcolor{lightblue}
        \hspace*{1em}\footnotesize{\textit{-- w/o OCR}} & 14.42 & 74.48 & 61.78 & 9.44  & 82.21 & 56.71 & 17.59 & 78.34 & 74.30 & 52.14 \\

        \rowcolor{lightblue}
        InternVL2.5-2B & 9.09 & 72.17 & 56.40 & 8.51  & 83.93 & 57.56 & 14.71 & 77.07 & 69.92 & 49.93 \\
        \rowcolor{lightblue}
        \hspace*{1em}\footnotesize{\textit{-- w/o OCR}} & 5.61  & 69.99 & 53.58 & 6.05  & 81.26 & 52.71 & 12.86 & 74.36 & 67.50 & 47.10\\

        \rowcolor{lighterblue}
        Qwen2.5-7B-Instruct & 10.62 & 73.79 & 58.65 & 15.64 & 86.98 & 62.43 & 21.12 & 80.42 & 75.99 & 53.96\\
        \rowcolor{lighterblue}
        Qwen2-7B-Instruct& 6.66 & 71.88 & 55.58 & 10.86 & 87.37 & 63.61 & 18.43 & 80.33 & 75.59 & 52.26 \\
        \rowcolor{lighterblue}
        InternLM2.5-7B-Chat& 2.70 & 67.25 & 49.98 & 8.97  & 85.92 & 59.92 & 4.38  & 73.09 & 64.98 & 46.35\\
        \rowcolor{lighterblue}
        Qwen2.5-3B-Instruct  & 10.20  & 72.46 & 56.81 & 12.83 & 85.27 & 60.34 & 17.83 & 78.25 & 72.69 & 51.85 \\
        \rowcolor{lighterblue}
        InternLM2.5-1.8B-Chat  & 0.61 & 60.11 & 40.22 & 1.39  & 80.50 & 49.35 & 1.21  & 64.21 & 52.62 & 38.91\\
    \midrule
        \multicolumn{10}{c}{\textbf{SFT with LoRA sample 1000}} \\
        \midrule
        \rowcolor{lightblue}
        Qwen2.5-VL-7B-LoRA & \textbf{26.94} & \textbf{81.27}  & \textbf{71.18}  & 18.73 & 88.24 & 64.99 & 26.98 & 82.96 & 79.58 & \textbf{60.10} \\
        \rowcolor{lightblue}
        \hspace*{1em}\footnotesize{\textit{-- w/ OCR}}  & 17.99 & 78.01 & 64.64 & 17.24 & 88.54 & 64.73 & 26.07 & 83.36 & 79.68 & 57.81 \\

        \rowcolor{lightblue}
        LLaVA-OneVision-7B-LoRA & 15.53 & 77.46 & 63.87 & 13.51 & 87.32 & 62.95 & 18.95 & 79.44 & 72.72 & 54.64 \\
        \rowcolor{lightblue}
        \hspace*{1em}\footnotesize{\textit{-- w/o OCR}} & 14.37 & 76.23 & 62.54 & 0.90  & 81.08 & 42.95 & 1.68  & 67.22 & 49.99 & 44.11 \\

        \rowcolor{lightblue}
        InternVL2.5-8B-LoRA & 14.85 & 76.59 & 62.01 & 16.64 & 88.57 & 64.92 & 22.84 & 81.65 & 76.46 & 56.06 \\
        \rowcolor{lightblue}
        \hspace*{1em}\footnotesize{\textit{-- w/o OCR}} & 15.00 & 76.00 & 61.88 & 13.96 & 87.41 & 62.48 & 19.87 & 79.53 & 73.37 & 54.39 \\

        \rowcolor{lightblue}
        Qwen2.5-VL-3B-LoRA  & 16.08 & 77.07 & 64.82 & 16.36 & 87.47 & 64.13 & 23.68 & 82.41 & 78.77 & 56.75 \\
        \rowcolor{lightblue}
        \hspace*{1em}\footnotesize{\textit{-- w/o OCR}} & 17.14 & 76.76 & 64.77 & 15.41 & 84.73 & 61.12 & 21.87 & 80.92 & 76.41 & 55.46 \\

        \rowcolor{lightblue}
        InternVL2.5-2B-LoRA & 10.86 & 73.85 & 56.56 & 12.08 & 87.41 & 61.76 & 16.96 & 78.54 & 70.10 & 52.01 \\
        \rowcolor{lightblue}
        \hspace*{1em}\footnotesize{\textit{-- w/o OCR}} & 6.30  & 69.71 & 50.38 & 7.30  & 85.33 & 56.78 & 13.69 & 75.78 & 66.77 & 48.00 \\

        \midrule
        \multicolumn{10}{c}{\textbf{SFT with LoRA sample 500}} \\
        \midrule
        
        \rowcolor{lightblue}
        Qwen2.5-VL-7B-LoRA  & 16.25 & 77.04 & 64.85 & 15.91 & 87.67 & 64.35 & 22.82 & 82.31 & 78.59 & 56.64 \\
        \rowcolor{lightblue}
        \hspace*{1em}\footnotesize{\textit{-- w/o OCR}} & 17.11 & 76.71 & 64.71 & 15.44 & 85.02 & 61.65 & 21.04 & 80.64 & 76.62 & 55.44 \\

        \rowcolor{lightblue}
        LLaVA-OneVision-7B-LoRA & 15.48 & 77.34 & 63.96 & 13.49 & 87.28 & 63.01 & 17.23 & 78.38 & 70.52 & 54.08 \\
        \rowcolor{lightblue}
        \hspace*{1em}\footnotesize{\textit{-- w/o OCR}} & 15.79 & 76.54 & 63.26 & 2.17  & 81.79 & 43.35 & 2.06  & 66.48 & 48.18 & 44.40 \\
        
        \rowcolor{lightblue}
        InternVL2.5-8B-LoRA & 14.07 & 76.17 & 61.36 & 16.56 & 88.58 & 65.03 & 23.34 & 81.52 & 76.60 & 55.91 \\
        \rowcolor{lightblue}
        \hspace*{1em}\footnotesize{\textit{-- w/o OCR}} & 9.63  & 74.52 & 59.99 & 12.62 & 87.19 & 62.45 & 18.61 & 78.89 & 72.72 & 52.96 \\

        \rowcolor{lightblue}
        Qwen2.5-VL-3B-LoRA  & 16.49 & 77.18 & 64.95 & 15.41 & 87.49 & 63.92 & 24.59 & 82.54 & 78.58 & 56.79 \\
        \rowcolor{lightblue}
        \hspace*{1em}\footnotesize{\textit{-- w/o OCR}} & 17.15 & 76.78 & 65.00 & 15.00 & 84.75 & 61.09 & 21.34 & 80.82 & 76.55 & 55.39 \\

        \rowcolor{lightblue}
        InternVL2.5-2B-LoRA & 9.27  & 72.72 & 54.74 & 11.61 & 87.21 & 61.20 & 12.92 & 77.00 & 67.89 & 50.51 \\
        \rowcolor{lightblue}
        \hspace*{1em}\footnotesize{\textit{-- w/o OCR}} & 9.71  & 74.43 & 59.90 & 13.01 & 87.35 & 62.39 & 18.52 & 78.77 & 72.55 & 52.96 \\
        
        \bottomrule
    \end{tabular}
    }
    \caption{Extended experimental results (Table~\ref{tab:eval_3dataset}) of different models (All Language Directions).}
    \label{tab:results}
\end{table*}

\end{document}